\documentclass[conference]{IEEEtran}

\usepackage{amsmath,amsfonts,bm,amsthm,mathtools}

\newcommand{\T}{^\mathrm{T}}

\def\eqref#1{equation~\ref{#1}}

\def\1{\bm{1}}

\newcommand{\Rb}{\mathbb{R}}

\newcommand{\Eb}{\mathbb{E}}

\newcommand{\ExP}[2]{\Eb_{{#1}}{\left[#2\right]}}

\def\rd{{\textnormal{d}}}

\def\calN{{\mathcal{N}}}
\def\calF{{\mathcal{F}}}

\DeclareMathAlphabet{\mathsfit}{\encodingdefault}{\sfdefault}{m}{sl}
\SetMathAlphabet{\mathsfit}{bold}{\encodingdefault}{\sfdefault}{bx}{n}

\newcommand{\logr}{\log_r}
\newcommand{\expr}{\exp_r}

\newcommand{\KL}[2]{\mathcal{D}_{KL}\left({#1}\parallel {#2}\right)}

\newcommand{\D}[3]{\mathcal{D}_{#1}\left({#2}\parallel {#3}\right)}
\newcommand{\Var}{\mathrm{Var}}

\newcommand{\Sr}{\mathcal{S}_r}

\DeclareMathOperator*{\argmax}{arg\,max}
\DeclareMathOperator*{\argmin}{arg\,min}
\DeclareMathOperator\erf{erf}

\DeclarePairedDelimiter\abs{\lvert}{\rvert}
\DeclarePairedDelimiter\norm{\lVert}{\rVert}

\DeclareMathOperator{\diag}{diag}

\newcommand{\wrapcell}[1]{\begin{tabular}[c]{@{}c@{}}#1\end{tabular}}

\usepackage[numbers]{natbib}
\usepackage{multicol}
\usepackage{xr-hyper}
\usepackage[bookmarks=true]{hyperref}
\usepackage{times}

\usepackage[nolist]{acronym}
\begin{acronym}
\acro{ARA}{Absolute Risk Aversion}
\acro{VI}{Variational Inference}
\acro{VI-MPC}{Variational Inference-Model Predictive Control}
\acro{SOC}{Stochastic Optimal Control}
\acro{SVGD}{Stein Variational Gradient Descent}
\acro{SS}{Stochastic Search}
\acro{MPPI}{Model Predictive Path Integral}
\acro{MPC}{Model Predictive Control}
\acro{CEM}{Cross Entropy Method}
\acro{RL}{Reinforcement Learning}
\acro{KL}{Kullback-Leibler}
\acro{EM}{Expectation Maximization}
\end{acronym}
\usepackage{hyperref}
\hypersetup{colorlinks = true,
            linkcolor = blue,
            urlcolor  = black,
            citecolor = blue,
            anchorcolor = blue}

\usepackage{url}

\usepackage[utf8]{inputenc} 
\usepackage[T1]{fontenc}    
\usepackage{url}            
\usepackage{booktabs}       
\usepackage{amsfonts}       
\usepackage{nicefrac}       
\usepackage{microtype}      
\usepackage{algorithm}
\usepackage{amsmath}
\usepackage{amssymb}
\usepackage{bm}
\usepackage{mathtools}
\usepackage{amsthm}
\usepackage{mathrsfs}
\usepackage{subcaption}
\usepackage{graphicx} 
\usepackage{caption}
\usepackage{algorithmic}
\usepackage[nolist]{acronym}
\usepackage{bbm}
\usepackage{xcolor}
\usepackage{wrapfig}
\usepackage{siunitx}
\usepackage{multirow}

\let\labelindent\relax
\usepackage{enumitem}

\usepackage[symbol]{footmisc}
\usepackage{float}
\usepackage[capitalise]{cleveref}
\usepackage{caption}

\usepackage{tikz}
\usepackage{etoolbox}
\usepackage{siunitx}
\usepackage{booktabs}
\usepackage{natbib}

\robustify\bfseries

\usepackage{ulem}

\makeatletter
\newcommand*{\addFileDependency}[1]{
  \typeout{(#1)}
  \@addtofilelist{#1}
  \IfFileExists{#1}{}{\typeout{No file #1.}}
}
\makeatother

\newcommand*{\myexternaldocument}[1]{%
    \externaldocument{#1}%
    \addFileDependency{#1.tex}%
    \addFileDependency{#1.aux}%
}

\myexternaldocument{paper/supplementary}

\begin{document}

\title{Variational Inference MPC using Tsallis Divergence}




%
\author{\authorblockN{Ziyi Wang\authorrefmark{1}\authorrefmark{2},
Oswin So\authorrefmark{1},
Jason Gibson, 
Bogdan Vlahov,
Manan S. Gandhi\\
Guan-Horng Liu and Evangelos A. Theodorou}
\authorblockA{Autonomous Control and Decision Systems Lab\\
Georgia Institute of Technology,
Atlanta, Georgia\\
\authorrefmark{1}These authors contributed equally\\
\authorrefmark{2}Correspondence to: \href{mailto:ZiyiWang@gatech.edu}{ZiyiWang@gatech.edu}}}

\maketitle

\begin{abstract}
In this paper, we provide a generalized framework for Variational Inference-Stochastic Optimal Control  by using the non-extensive Tsallis divergence. By incorporating the deformed exponential function into the optimality likelihood function, a novel Tsallis Variational Inference-Model Predictive Control algorithm is derived, which includes prior works such as Variational Inference-Model Predictive Control, Model Predictive Path Integral Control, Cross Entropy Method, and Stein Variational Inference Model Predictive Control as special cases.  The proposed algorithm  allows for effective control of the cost/reward transform  and is characterized by superior performance in terms of mean and  variance reduction of the associated cost. The aforementioned features are supported by a theoretical and numerical analysis on the level of risk sensitivity of the proposed algorithm as well as simulation experiments on 5 different robotic systems with 3 different policy parameterizations. 
\end{abstract}

\IEEEpeerreviewmaketitle

\section{Introduction}
\label{sec:intro}



\ac{VI} is a powerful tool for approximating the posterior distribution of the unobserved random variables \cite{blei2017variational}. \ac{VI} recasts the approximation problem as an optimization problem. Instead of directly approximating the target distribution $p(z|x)$ of the latent variable $z$, \ac{VI} minimizes the \ac{KL} divergence between a tractable variational distribution $q(z)$ and the target distribution. Due to its faster convergence and comparable performance to Markov Chain Monte Carlo sampling methods, \ac{VI} has received increasing attention in machine learning and robotics \cite{zhang2018advances, shankar2020learning, pignat2020variational}.

\ac{VI} has been applied to \ac{SOC} problems recently. In \citet{okada2020variational}, the authors formulated the \ac{SOC} problem as a \ac{VI} problem by setting the desired policy distribution as the target distribution. The \ac{VI}-\ac{SOC} framework works directly in the space of policy distributions instead of specific policy parameterizations in most \ac{SOC} and \ac{RL} frameworks. This gives rise to a unified derivation for a variety of parametric policy distributions, such as unimodal Gaussian and Gaussian mixture in \citet{okada2020variational}. \citet{lambert2020stein} derived the \ac{VI}-\ac{SOC} algorithm for non-parametric policy distribution using Stein variational gradient descent. Apart from the policy distribution, the \ac{VI}-\ac{SOC} framework is characterized by two components, the optimality likelihood function and distributional distance metric. The optimality likelihood function measures the likelihood of trajectory samples to  be optimal and defines the cost/reward transform to allow for different algorithmic developments. The distributional distance metric is the measure of distance between the variational distribution and target distribution.

In existing \ac{VI}-\ac{SOC} works, the \ac{KL} divergence is used as the distributional distance metric due to its simplicity. On the other hand, recent advances in \ac{VI} research involve extending the framework to other statistical divergences, such as the $\alpha$-divergence \cite{li2016r} and $\chi$-divergence \cite{dieng2016variational}. In \citet{wang2018variational, regli2018alpha}, the authors proposed variants of the $\alpha$-divergence to improve the performance and robustness of the inference algorithm. \citet{wan2020f} further extended the \ac{VI} framework to $f$-divergence, which is a broad statistical divergence family that recovers the \ac{KL}, $\alpha$ and $\chi$-divergence as special cases.

Tsallis divergence is another generalized divergence rooted in non-extensive statistical mechanics \cite{tsallis1988possible}. Tsallis divergence and Tsallis entropy are generalizations of the \ac{KL} divergence and Shannon entropy respectively. The Tsallis entropy is non-additive (hence non-extensive) in the sense that for a system composed of (probabilistically) independent subsystems, the total entropy differs from the sum of the entropies of the subsystems \cite{tsallis2005asymptotically}. Over the last two decades, an increasing number of complex natural, artificial and social systems have verified the predictions and consequences derived from the Tsallis entropy and divergence \cite{lutz2003anomalous, liu2008superdiffusion, devoe2009power, lee2019tsallis}. \citet{lee2019tsallis} demonstrated that Tsallis entropy regularization leads to improved performance and faster convergence in \ac{RL} applications.

In this paper, we provide a generalized formulation of the \ac{VI}-\ac{SOC} framework using the Tsallis divergence and introduce a novel \ac{MPC} algorithm. The main contribution of our work is threefold:
\begin{itemize}
    \item We propose the Tsallis \ac{VI}-\ac{MPC} algorithm, which allows for additional control of the shape of the cost transform compared to previous \ac{VI}-\ac{MPC} algorithms using \ac{KL} divergence.
    \item We provide a holistic view of connections between Tsallis \ac{VI}-\ac{SOC} and the state-of-the-art \ac{MPPI} control, \ac{CEM} and \ac{SS} methods.
    \item We show, both analytically and numerically, that the proposed Tsallis \ac{VI}-\ac{SOC} framework achieves lower cost variance than \ac{MPPI} and \ac{CEM}. We further demonstrate the superior performance of Tsallis \ac{VI}-\ac{SOC} in mean cost and variance minimization on simulated systems from control theory and robotics for 3 different choices of policy distributions.
\end{itemize}

The rest of this paper is organized as follows: in \cref{sec:derivation}, we review \ac{KL} \ac{VI}-\ac{SOC} and derive the Tsallis \ac{VI}-\ac{SOC} framework. In \cref{sec:connections}, we discuss the connections between Tsallis \ac{VI}-\ac{SOC} and related works. A reparameterization and analysis of the framework is included in \cref{sec:analysis}, and we propose the novel Tsallis \ac{VI}-\ac{MPC} algorithm in \cref{sec:mpc_alg}. We showcase the performance of the proposed algorithm against related works in \cref{sec:simulation} and conclude the paper in \cref{sec:conclusion}.

\section{Tsallis Variational Inference-Stochastic Optimal Control}
\label{sec:derivation}
In this section, we review the \ac{VI}-\ac{SOC} formulation using KL divergence \cite{okada2020variational} and derive the novel Tsallis divergence \ac{VI}-\ac{SOC} framework.

\subsection{SOC Problem Formulation}
In discrete time \ac{SOC}, we work with trajectories $\tau\coloneqq\{X,U\}$ of state, $X\coloneqq\{x_0, x_1,\dots, x_T\}\in\Rb^{n_x\times {T+1}}$, and control, $U\coloneqq\{u_0, u_1,\dots, u_{T-1}\}\in\Rb^{n_u\times T}$ over a finite time horizon $T>0$.
The goal in \ac{SOC} problems is to minimize the expected cost defined by an arbitrary cost function $J:\Rb^{n_x\times T}\times \Rb^{n_u\times T-1}\rightarrow \Rb^+$ with initial state distribution $p(x_0)$ and state transition probability $p(x_{t+1}|x_t, u_t)$ corresponding to stochastic dynamics $x_{t+1}=F(x_t,u_t,\epsilon_t)$ where $\epsilon_t\in\Rb^{n_\epsilon}$ is the system stochasticity.

\subsection{KL VI-SOC}
\label{sec:kl_vi}

We can formulate the \ac{SOC} problem as an inference problem and apply methods from \ac{VI}. To apply \ac{VI} to \ac{SOC}, we introduce a dummy optimality variable $o\in\{0,1\}$ with $o=1$ indicating that the trajectory $\tau=\{X,U\}$ is optimal.
\cref{tab:VI_SOC_comparison} compares the differences in notation between conventional \ac{VI} methods and \ac{SOC} formulated as a \ac{VI} problem.

The objective of \ac{VI}-\ac{SOC} is to sample from the posterior distribution
\begin{equation}
\begin{split}
p(\tau|o) &= \frac{p(o=1|\tau) p(\tau)}{p(o)}\\
&= \frac{p(o=1|\tau)}{p(o=1)}p(x_0)\prod_{t=0}^{T-1}p(x_{t+1}|x_t,u_t)p(u_t).
\end{split}
\end{equation}
Here $p(x_{t+1}|x_t,u_t)$ is the state transition probability, $p(x_0)$ is the initial state distribution and $p(u_t)$ is some prior control distribution (e.g. zero mean Gaussian or uniform distribution).
For simplicity, we use $o$ to indicate $o=1$ and $o'$ for $o=0$ from here on.
We can now formulate the \ac{VI}-\ac{SOC} objective as minimizing the distance between a controlled distribution $q(\tau)$ and the target distribution $p(\tau|o)$
\begin{equation}
\begin{split}
q^*(\tau)
&= \argmin_{q(\tau)} \KL{q(\tau)}{p(\tau|o)} \\
&= \argmin_{q(\tau)} \Eb_{q(\tau)}\Bigg[\log\frac{ p(X | U) q(U)}{ \frac{p(o|\tau)}{p(o)} p(X | U) p(U)}\Bigg] \\
&= \argmin_{q(\tau)} \Eb_{q(\tau)}\Bigg[-\log p(o|\tau) + \sum_{t=0}^{T-1}\log \frac{q(u_t)}{p(u_t)} \Bigg],
\end{split}
\end{equation}
where the constant $p(o)$ is dropped. The first term in the objective measures the likelihood of a trajectory being optimal while the second term serves as regularization.
Splitting the expectation in the first term, we get
\begin{equation}
\begin{split}\label{eq:kl_vi_soc:kl_vi_obj}
q^*(U) = \argmin_{q(U)} & \Big\{ -\Eb_{q(U)}[\Eb_{p{(X|U)}}[\log p(o|\tau)]] \\&+ \KL{q(U)}{p(U)} \Big\},
\end{split}
\end{equation}
where $p(U)$ and $q(U)$ represent $\prod_{t=0}^{T-1} p(u_t)$ and $\prod_{t=0}^{T-1} q(u_t)$ respectively due to independence and $p(X|U)=p(x_0)\prod_{t=0}^{T-1}p(x_{t+1}|x_t,u_t)$.

\textbf{Optimality Likelihood:}
The optimality likelihood in \cref{eq:kl_vi_soc:kl_vi_obj} can be parameterized by a non-increasing function of the trajectory cost $J(X, U)$ as $ p(o|\tau) \coloneqq f(J(X,U))$.
The monotonicity requirement ensures that trajectories incurring higher costs are always less likely to be optimal.
Common choices of $f$ include $f(x)=\exp(-x)$ and $f(x)=\bm{1}_{ \{x\leq \gamma\} }$.
In this paper, we choose $f(x)=\exp(-x)$ such that $\log p(o|\tau) = -J(X, U)$.
To avoid excessive notation, we abuse the notation to define 
$J(U) \coloneqq \Eb_{p(X|U)}[J(X, U)]$ and the $N$-sample empirical approximation of the expectation $J=\sum_{n=1}^N [J^{(n)}]$.
Hence, \cref{eq:kl_vi_soc:kl_vi_obj} takes the form
\begin{equation}\label{eq:kl_vi_soc:kl_vi_obj2}
q^*(U) = \argmin_{q(U)} \Eb_{q(U)}[J(X, U)] + \KL{q(U)}{p(U)},
\end{equation}
which can be interpreted as an application of the famous maximum entropy principle.

\begin{table}[t]
    \centering
    \caption{Notation comparison between Variational Inference and Variational Inference-Stochastic Optimal Control.}
    \begin{tabular}{@{} l l l l @{}}
    \toprule
    \multicolumn{2}{c}{\ac{VI}} & \multicolumn{2}{c}{\ac{VI}-\ac{SOC}} \\ \cmidrule(r){1-2} \cmidrule(lr){3-4}
    {Notation} & {Meaning} & {Notation} & {Meaning}\\
    \midrule
    $x$ & data & $o$ & optimality \\
    \addlinespace[0.2em]
    $z$ & latent variable & $\tau$ & trajectory \\
    \addlinespace[0.2em]
    $p(z)$ & prior distribution & $p(\tau)$ &  prior distribution \\
    \addlinespace[0.2em]
    $q(z)$ & variational distribution & $q(\tau)$ & controlled distribution \\
    \addlinespace[0.2em]
    $p(x|z)$ & generative model & $p(o|\tau)$ & optimality likelihood \\
    \addlinespace[0.2em]
    $p(z|x)$ & posterior distribution & $p(\tau|o)$ & optimal distribution \\
    \bottomrule
    \end{tabular}
    \label{tab:VI_SOC_comparison}
\end{table}

\subsection{Tsallis VI-SOC}
\label{sec:tsallis_vi}
In this subsection, we use the Tsallis divergence as the regularization function and derive the Tsallis-\ac{VI}-\ac{SOC} algorithm.
First, we define the deformed logarithm and exponential as
\begin{align}
\logr(x) & = \frac{x^{r-1}-1}{r-1} \label{eq:tsallis_vi:logr}, \\
\expr(x) & = (1+(r-1)x)_+^{\frac{1}{r-1}}, \label{eq:tsallis_vi:expr}
\end{align}
where $(\cdot)_+\coloneqq\max(0, \cdot)$ and $r>0$.
Using $\logr$ and $\expr$ we can define the Tsallis entropy and the corresponding Tsallis divergence \cite{tsallis1988possible} as
\begin{equation}
\begin{split}
\Sr(q(z)) \coloneqq & -\Eb_q[\logr q(z)] \\
= & -\frac{1}{r-1}\left(\int q(z)^r \rd z - 1\right),
\end{split}
\end{equation}
and
\begin{equation}
\begin{split}
\D{r}{q(z)}{p(z)} \coloneqq & \Eb_q\left[\logr\frac{q(z)}{p(z)}\right] \\
= &\frac{1}{r-1}\left(\int q(z)\left(\frac{q(z)}{p(z)}\right)^{r-1}\rd z- 1\right).
\end{split}
\end{equation}
Note that as $r\rightarrow 1$, $\logr\rightarrow \log$, $\expr\rightarrow \exp$, $\mathcal{D}_r\rightarrow\mathcal{D}_{KL}$ and $\mathcal{S}_r\rightarrow \mathcal{S}$, where $\mathcal{S}$ is the Shannon entropy, recovering the KL \ac{VI}-\ac{SOC} framework.

\textbf{Versions of Tsallis Statistics:}
In our definition of the deformed logarithm and exponential in \cref{eq:tsallis_vi:logr,eq:tsallis_vi:expr}, we use the variable $r$ instead of the $q$ used in most literature to avoid assigning multiple meanings to $q$.
Also, our definition of $\logr$ and $\expr$ differ from its original definitions \cite{tsallis1994numbers}, but
the original can be recovered with $r' = 2 - r$, where $r'$ corresponding to the value used in \cite{tsallis1994numbers}.
Additionally, there are multiple formulations of the Tsallis entropy differing in their definitions of the internal energy and how expectations are taken \cite{tsallis1988possible, curado1991generalized, tsallis1998role}.
We have chosen to use the formulation from \cite{tsallis1988possible} due to its simplicity in computing the expectation.
However, as shown in \cite{ferri2005equivalence}, they are equivalent and can be recovered from each other via a change of variables.

We can now define a new objective by replacing the KL divergence regularizer in \cref{eq:kl_vi_soc:kl_vi_obj2} with the Tsallis divergence and introducing a parameter $\lambda$ that multiplies the optimality likelihood to make the regularization strength tunable:
\begin{equation}\label{eq:tsallis_vi_soc:tsallis_vi_obj}
    q^*(U) = \argmin_{q(U)} \lambda^{-1} \Eb_{q(U)}[J] + \D{r}{q(U)}{p(U)}.
\end{equation}
The controlled distribution has to satisfy the additional constraint of $\int q(U)\, \rd U = 1$.
The optimal policy distribution $q^*$ can be explicitly solved for with the form
\begin{equation}\label{eq:tsallis_vi_soc:dist_update}
q^*(U)
=\frac{\expr \left( -\tilde{\lambda}^{-1} J\right)p(U)} {\int \expr \left( -\tilde{\lambda}^{-1} J\right)p(U)\, \rd U},
\end{equation}
where $\tilde{\lambda} = \alpha (r-1) \lambda$ with $\alpha$ being the Lagrange multiplier for the constraint that $q^*$ integrates to $1$.
A detailed derivation can be found in \cref{SM:optimal_dist_derivation} of the supplementary materials. We can now use \cref{eq:tsallis_vi_soc:dist_update} to obtain the optimal control distribution by transforming the prior distribution. 


\subsection{Update Laws} \label{sec:update_law}
In general, it is computationally inefficient to sample from $q^*(U)$ via \cref{eq:tsallis_vi_soc:dist_update} directly.
Instead, we can approximate $q^*(U)$ by some policy $\pi(U)$ lying in a class $\Pi$ of tractable distributions and
solve for an iterative update law by minimizing the KL divergence between $\pi(U)$ and $q^*(U)$:
\begin{equation} \label{eq:update_laws:obj_fn}
    \pi^{k+1}(U) = \argmin_{\pi(U) \in \Pi} \KL{q^*(U)}{\pi(U)} .
\end{equation}
However, because one can only evaluate $q^*(U)$ at a finite number of points $\{ U^{(n)} \}_{n=1}^N$, we instead approximate $q^*(U)$ by the empirical distribution $\tilde{q}^*(U)$ with weights $w^{(n)}$:
\begin{align}
    \tilde{q}^*(U) = \sum_{n=1}^N w^{(n)} \bm{1}_{\{U = U^{(n)} \}}, \\
    w^{(n)} = \frac{q^*(U^{(n)})}{\sum_{n'=1}^N q^*(U^{(n')}) } .
\end{align}

We now solve \cref{eq:update_laws:obj_fn} for 3 different policy classes: unimodal Gaussian, Gaussian mixture and a nonparametric policy corresponding to \ac{SVGD} \cite{lambert2020stein, liu2016stein}.
The full derivations for each can be found in the supplementary material in \cref{SM:sec:derive_update_laws}.

\textbf{Unimodal Gaussian:} For a unimodal Gaussian policy distribution with parameters $\Theta\coloneqq\{ (\mu_t, \Sigma_t) \}_{t=0}^{T-1}$, the update laws for the $k+1$th iteration take the form of
\begin{align} 
\mu^{k+1}_t &=
    \sum_{n=1}^N w^{(n)} u_t^{(n)}, \label{eq:mean_update} \\
\Sigma^{k+1}_t &=
    \sum_{n=1}^N w^{(n)} (u_t^{(n)}-\mu_t^{k+1})(u_t^{(n)}-\mu_t^{k+1})\T \label{eq:var_update} .
\end{align}

\textbf{Gaussian Mixture:} Alternatively, the policy distribution can be an $L$-mode mixture of Gaussian distribution with parameters $\Theta \coloneqq \{ \theta_l \}_{l=1}^{L}$ for $\theta_l \coloneqq (\phi_l, \{ \mu_{l, t}, \Sigma_{l, t} \}_{t=0}^{T-1})$, where $\phi_l$ is the mixture weight for the $l$th component.
Although it is not possible to directly solve \cref{eq:update_laws:obj_fn} in this case, we draw from \ac{EM} to derive an iterative update scheme for the $k+1$th iteration with the form
\begin{align}
\phi_l^{k+1} &= \frac{N_l}{\sum_{l'=1}^L N_{l'}},\\
\mu_{l,t}^{k+1} &= \frac{1}{N_l}\sum_{n=1}^N \eta_l(u_t^n) w^{(n)} u_t^n,\\
\Sigma_{l,t}^{k+1} &= \frac{1}{N_l}\sum_{n=1}^N \eta_l(u_t^n) w^{(n)} (u^n_t-\mu_{l,t}^{k+1})(u^n_t-\mu_{l,t}^{k+1})\T,
\end{align}
where
\begin{align}
\eta_l(u_t^{(n)}) &= \frac{\phi_l^k\calN(u_t^{(n)}|\mu_{l,t}^k, \Sigma_{l,t}^k)}{\sum_{l'=1}^L \phi_{l'}^k \calN(u_t^{(n)} | \mu_{l',t}^k, \Sigma_{l',t}^k)},\\
N_{l} &= \sum_{n=1}^N \sum_{t=0}^{T-1} \eta_l(u_t^{(n)}) w^{(n)}.
\end{align}

\textbf{Stein Variational Policy:} The policy can also be a non-parametric distribution approximated by a set of particles $\Theta\coloneqq\{\theta_l\}_{l=1}^L$ for some parametrized policy $\hat{\pi}(U; \theta)$. In \cite{lambert2020stein}, $\hat{\pi}$ is taken to be a unimodal Gaussian with fixed variance, where $\theta \in \Rb^{n_x \times (T-1)}$ corresponds to the mean.
The update law of each Stein particle for the $k+1$th iteration has the form
\begin{align}
    \theta_l^{k+1}
        &= \theta_l^{k} + \epsilon \hat{\phi}^*(\theta_l^k), \\
    \hat{\phi}^*(\theta)
        &= \sum_{l=1}^L \hat{k}(\theta_{l}, \theta) G(\theta_m) + \nabla_{\theta_l} \hat{k}(\theta_l, \theta), \\
    G(\theta_l) &= \frac{\sum_{s=1}^S w^{(l, s)} \nabla_\theta \log \hat{\pi}(U^{(n)}, \theta_{l})}{\sum_{s=1}^S w^{(l, s)}} ,
\end{align}
where the $N$ is chosen such that $N = LS$, $\hat{k}$ is a kernel function, and $w^{(l, s)} = w^{(m+L(l-1))}$, where $S$ denotes the number of rollouts for each of the $L$ particle.
As noted in \cite{zhuo2018message}, \ac{SVGD} becomes less effective as the dimensionality of the particles increases due to the inverse relationship between the repulsion force in the update law and the dimensionality.
Hence, we follow \cite{lambert2020stein} in choosing a sum of local kernel functions as our choice of $\hat{k}$.

\bgroup
\begin{table*}[t]
    \centering
    \caption{Comparison of the objective and the update law for a unimodal Gaussian policy with fixed variance between different SOC approaches.}
    \begin{tabular}{@{} l l l @{}}
    \toprule
    \multicolumn{1}{c}{Formulation} & \multicolumn{1}{c}{Objective (Minimize)} & \multicolumn{1}{c}{Update Law for Unimodal Gaussian} \\
    \midrule
     Tsallis VI-SOC & $-\lambda\Eb[\log p(o|\tau)]+\D{r}{q}{p}$ & $\theta_t^{k+1} = \sum_{n=1}^{N}\frac{\expr \left(-\tilde{\lambda}^{-1}J^n\right)s(u^n_t) u^n_t }{\sum_{n'=1}^N \expr \left(-\tilde{\lambda}^{-1}J^{n'}\right)s(u^{n'}_t)}$ \\
    \addlinespace[0.5em]
    KL VI-SOC & $-\lambda\Eb[\log p(o|\tau)]+\KL{q}{p}$ & $\theta_t^{k+1} = \sum_{n=1}^{N}\frac{\exp \left(-\lambda^{-1}J^n\right)s(u^n_t) u^n_t }{\sum_{n'=1}^N \exp \left(-\lambda^{-1}J^{n'}\right)s(u^{n'}_t)}$ \\
    \addlinespace[0.5em]
    SS-SOC & $\Eb[S(J)]$ & $\theta_t^{k+1} = \theta_t^k + \beta\sum_{n=1}^N\frac{S(J^n)  (u_t^n - \frac{1}{N}\sum_{\tilde{n}=1}^N u_t^{\tilde{n}}) }{\sum_{n'=1}^N S(J^{n'})  u_t^{n'} }$ \\
    \addlinespace[0.5em]
    MPPI & $\Eb[J] + \KL{q}{p}$ & $\theta_t^{k+1} = \theta_t^k + \sum_{n=1}^N\frac{\exp(-\lambda^{-1} J^n)s(u_t^n)u_t^n}{\sum_{n'=1}^N \exp(-\lambda^{-1} J^{n'})s(u_t^{n'})}$ \\
    \addlinespace[0.5em]
    CEM & $\Eb[J]$ & $\theta_t^{k+1}=\alpha \theta_t^k + (1-\alpha) \sum_{n=1}^N \frac{\bm{1}_{ \{J^n\leq \gamma\} } u_t^n }{\sum_{n'=1}^N \bm{1}_{ \{J^{n'}\leq \gamma\} }}$ \\
    \bottomrule
    \end{tabular}
    \label{tab:connection_comparison}
\end{table*}
\egroup

\section{Connections to Related Works}
\label{sec:connections}

In this section, we compare \ac{VI}-\ac{SOC} with three sampling-based methods from optimization, thermodynamics, and information theory that have been applied to stochastic control problems.
A comparison of problem formulations and update laws for different approaches with a unimodal Gaussian policy is in \Cref{tab:connection_comparison}.

\subsection{Stochastic Search}
\label{sec:connection_to_ss}
\ac{SS} \cite{zhou2014gradient} is a stochastic optimization scheme and has also been applied to the \ac{SOC} setting \cite{boutselis2020constrained, wang2020adaptive}. \ac{SS}-\ac{SOC} parameterizes the control policy with a distribution from the exponential family during problem formulation and optimizes with respect to policy parameters whereas \ac{VI}-\ac{SOC} performs optimization at the distribution level.
The \ac{SS}-\ac{SOC} framework formulates the problem as
\begin{equation}\label{eq:ss_objective}
\theta^* = \argmax_\theta \Eb_{\pi(U;\theta)}\left[S( \Eb_{p(X|U)} [J(X, U)] )\right],
\end{equation}
where $S(\cdot)$ is a monotonically decreasing shape function.
Note that in \cref{eq:ss_objective}, the expectation taken with respect to $p(X|U)$ is inside the shape function $S$ as opposed to outside as in \cref{eq:kl_vi_soc:kl_vi_obj}.
For convex shape/optimality likelihood functions, the objective in VI-SOC corresponds to an upper bound of that in SS-SOC, i.e.,
\begin{equation}
    S(\Eb_{p(X|U)}[J(X, U)]) \leq \Eb_{p(X|U)}[S(J(X, U))].
\end{equation}
A detailed comparison of the two formulations can be found in \cite{okada2020variational}.

The parameter update of SS-SOC has the form
\begin{equation}\label{eq:ss_update_law}
\theta_t^{k+1} =
\theta_t^k + \beta\sum_{n=1}^N\frac{S(J^{(n)})(T(u_t^{(n)})-\frac{1}{N}\sum_{\tilde{n}=1}^NT(u^{(\tilde{n})}_t))}{\sum_{n'=1}^N S(J^{(n')})},
\end{equation}
where $T(\cdot)$ denotes the sufficient statistic for the corresponding parameter and $\beta$ is the step size.
In the case of a unimodal Gaussian policy, the update in \cref{eq:ss_update_law} is equivalent to \cref{eq:mean_update,eq:var_update} with $S(J)=\expr(-\tilde{\lambda}^{-1}J)$.

\subsection{Cross Entropy Method}
\label{sec:connection_to_cem}
\ac{CEM} \cite{de2005tutorial} is a widely used algorithm in reinforcement learning and optimal control problems \cite{mannor2003cross, kobilarov2012cross}.
The objective function of \ac{CEM} minimizes the expected cost $\Eb [J]$. The policy update law for \ac{CEM} corresponds to that of \ac{SS}-\ac{SOC} in \cref{eq:ss_update_law} with shape function $S(J)=\bm{1}_{ \{J\leq \gamma\} }$ where $\gamma$ is the elite threshold.
As will be shown in \cref{sec:reparameterization}, \ac{CEM} is also a special case of the reparameterized Tsallis \ac{VI}-\ac{SOC} with $r\rightarrow \infty$.

\subsection{Model Predictive Path Integral Control}
\label{sec:connection_to_pi}
\ac{MPPI} is another approach closely related to \ac{VI}-\ac{SOC} and \ac{SS}-\ac{SOC} \cite{williams2018information, wang2019information}. The framework solves for the controls by minimizing the KL divergence between a controlled distribution and the optimal control distribution
\begin{equation}
q^*(U) = \frac{\exp\left(-\lambda^{-1} J  \right) p(U) }{\int \exp\left(-\lambda^{-1} J \right)  p(U) \; \rd U}.
\end{equation}
The optimal distribution achieves the free energy lower bound, $\calF=-\lambda \log\Eb_{p(U)}[\exp(-\frac{1}{\lambda} J)]$, which has been shown to be the solution of the Hamilton-Jacobi-Bellman equation \cite{theodorou2012relative}.
The optimal distribution here is equivalent to \cref{eq:tsallis_vi_soc:dist_update} when $r \to 1$. The corresponding update law can also be obtained from the \ac{SS}-\ac{SOC} framework with $S(J)=\exp(-\lambda^{-1}J)$.

\section{Analysis}
\label{sec:analysis}

\begin{figure}[t]
    \centering
    \includegraphics[width=\linewidth]{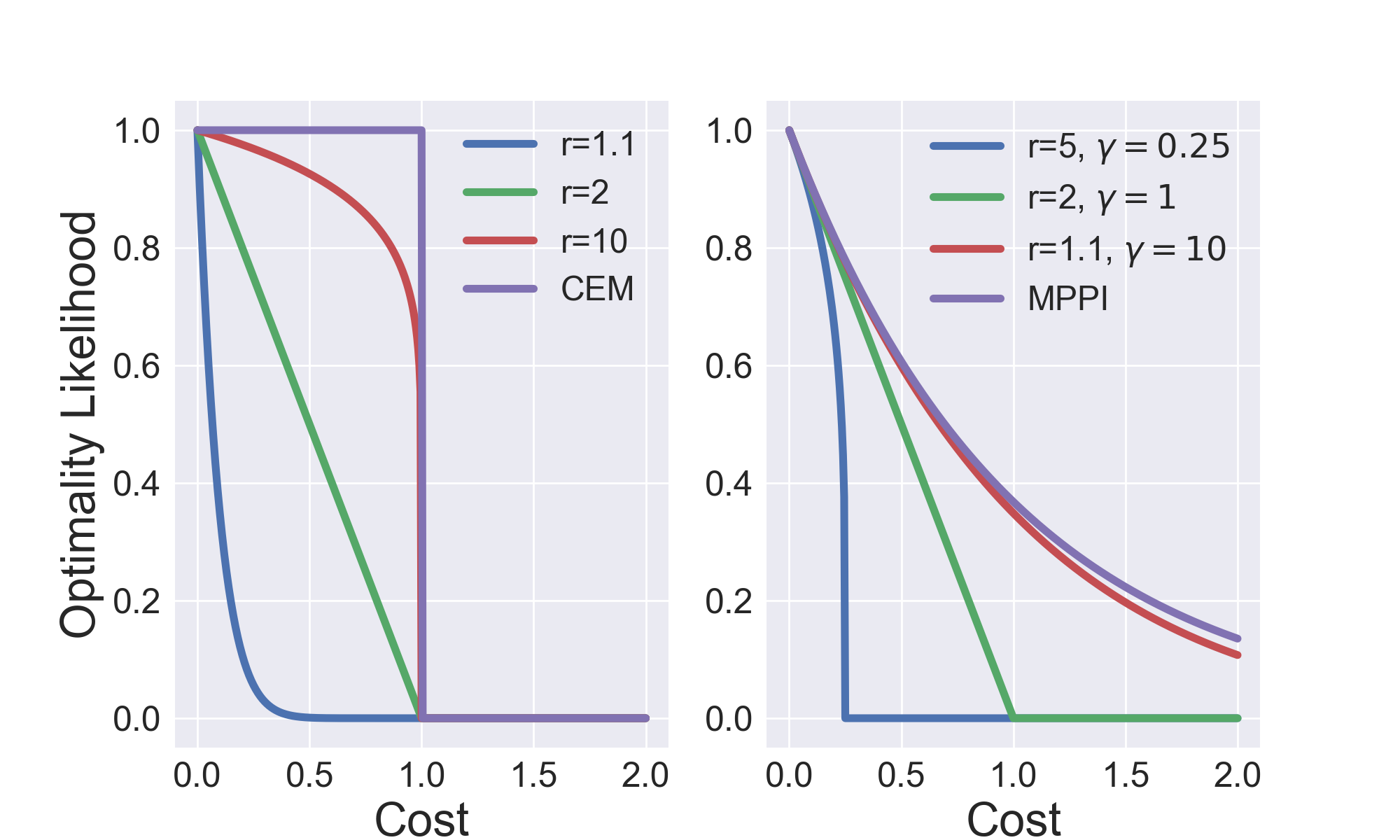}
    \caption{Reparameterized $\expr$ $(\cref{eq:reparameterization})$. \textit{Left}: $r$ is varied with fixed $\gamma=1$. It is clear that $\expr\to$ CEM as $r\to \infty$. \textit{Right}: $\gamma$ is adjusted accordingly such that $\expr\to$ MPPI as $r\to 1$.}
    \label{fig:exp_r}
\end{figure}

\subsection{Effect of \texorpdfstring{$\expr$}{exp r}}
\label{sec:reparameterization}
To facilitate the analysis, we focus on the $\expr$ term in \cref{eq:tsallis_vi_soc:dist_update}.
Reparametrizing $\tilde{\lambda}=(r-1)\gamma$, we get
\begin{equation}
\begin{split}\label{eq:reparameterization}
\expr(-\tilde{\lambda}^{-1} J) &= \left(1-\frac{J}{\gamma}\right)_+^{\frac{1}{r-1}} \\
&= \begin{cases}
        \exp\left(\frac{1}{r-1}\log \left(1-\frac{J}{\gamma}\right)\right), \, J<\gamma \\
        0, \hspace{9.6 em} J \geq \gamma
\end{cases},
\end{split}
\end{equation}
where $\gamma$ is now the threshold beyond which the optimality weight is set to 0.
The reparameterization adjusts the original parameter $\tilde{\lambda}$ at every iteration to maintain the same threshold $\gamma$, which is a more intuitive parameter.
In addition, we have observed that the reparameterized framework is easier to tune and achieves better performance than the original formulation. Therefore, we focus our analysis and simulations on only the reparameterized version hereon after. Note that in practice, it is easier to define an \textit{elite fraction}, which adjusts $\gamma$ based on the scale of the costs, instead of using $\gamma$ for easier tuning.

\Cref{fig:exp_r} illustrates the shapes of the function corresponding to different $r$ and $\gamma$ values.
For $J < \gamma$, as $r \rightarrow \infty$, $\expr(-\tilde{\lambda}^{-1} J) \rightarrow 1$.
Hence, $\expr(-\tilde{\lambda}^{-1} J)$ converges pointwise to the step function $\bm{1}_{ \{J \leq \gamma\} }$ with $r \to \infty$.
On the other hand, for any $0 < \frac{J}{\gamma} < 1$, we have that
\begin{equation}
    \frac{\expr(-\tilde{\lambda}^{-1} 0)}{\expr(-\tilde{\lambda}^{-1} J)} = \exp \left(-\frac{1}{r-1} \log \left(1-\frac{J}{\gamma} \right) \right),
\end{equation}
which tends to 0 as $r \to 1$.
Hence, $\exp\left(\frac{1}{r-1}\log \left(1-\frac{J}{\gamma}\right)\right)$ converges to $\bm{1}_{ \{J = 0 \}}$ as $r \to 1$. 

\subsection{Variance Reduction}
\label{sec:var_reduction}
With the connection to \ac{SS}, the effect of Tsallis divergence can be analyzed through the equivalent problem formulation in optimization.
In \Cref{sec:connection_to_ss}, it is shown that Tsallis \ac{VI}-\ac{SOC} corresponds to an upper bound of \ac{SS}-\ac{SOC} with objective 
\begin{equation}\label{eq:Tsallis_optimization_objective}
\theta^* = \argmax_\theta \Eb_{\pi(U;\theta)} \left[ \expr(-\tilde{\lambda}^{-1} J) \right].
\end{equation}

\renewcommand{\thefootnote}{\arabic{footnote}}

The variance reduction effect can be analyzed through the coefficient of \ac{ARA} \cite{pratt1978risk}
\begin{equation}
    A(J) = -\frac{S''(J)}{S'(J)},
\end{equation}
for the optimization problem
\begin{equation}
    \min_{\theta} \Eb_{p(X|U), \pi(U; \theta)} [S(J(X, U))],
\end{equation}
where Tsallis \ac{VI}-\ac{SOC} corresponds to 
$S(J)=\expr(-\tilde{\lambda}^{-1}J)$).
The \ac{ARA} coefficient measures a scaled ratio of terms corresponding to the mean and variance terms in the Taylor series expansion of the objective.
Negative value of the coefficient corresponds to a risk-averse objective, and positive value of the coefficient corresponds to risk-seeking behavior.\footnotemark
The \ac{ARA} coefficient of Tsallis \ac{VI}-\ac{SOC} is
\begin{equation}
A_{\text{Tsallis}}(J) = -\frac{r-2}{(r-1)(\gamma-J)}.
\end{equation}

The \ac{ARA} coefficients for \ac{MPPI} and \ac{CEM} are
\begin{align}
A_{\text{MPPI}}(J) &= \frac{1}{\lambda},\\
A_{\text{CEM}}(J) &= \lim_{k\rightarrow \infty} -k\tanh\left(\frac{1}{2}k(\gamma-J)\right).
\end{align}

Since $A_{\text{MPPI}}$ is a positive constant, it corresponds to a risk-seeking algorithm. For a cost below the elite threshold, $J<\gamma$, $A_{\text{CEM}}=-\infty$ and $A_{\text{Tsallis}} \lessgtr 0$ for $r\lessgtr 2$. This leads to the Tsallis \ac{VI}-\ac{SOC} framework achieving lower variance than \ac{MPPI}. For the same elite threshold $\gamma$ and a properly selected $r$, we hypothesize that Tsallis \ac{VI}-\ac{SOC} results in lower mean cost than \ac{CEM} since \ac{CEM} penalizes variance infinitely harder than the mean and assigns equal weights to all elite samples. A more detailed analysis and the derivation of \ac{ARA} coefficient are included in \cref{SM:variance_analysis} of the supplementary material.

\footnotetext{\ac{ARA} is originally used for utility maximization, where a larger ARA corresponds to greater risk aversion. This is the opposite in our case when we are performing cost minimization.}



\section{Model Predictive Control Algorithm}
\label{sec:mpc_alg}

\begin{algorithm}[t]
\caption{Tsallis Variational Inference MPC}
\begin{algorithmic}[1]
\STATE \textbf{Given:} $p(x_0)$: initial state distribution; $N$: number of policy samples; $M$: number of state  samples; $T$: MPC horizon; $T'$: number of MPC steps; $K$: optimization iterations per MPC step

\STATE Initialize $\Theta^0_0$
\STATE $\{x_0^{n,m}\}_{n,m=1}^{N,M} \sim p(x_0)$
\FOR{$t'=0$ to $T'-1$}
\FOR{$k=0$ to $K$}
\FOR{$n=1$ to $N$ \textit{in parallel}}
\FOR{$m=1$ to $M$ \textit{in parallel}}
\STATE $X^{n,m}, U^n=f_{\text{rollout}}(x_0^{n,m}, \Theta^k_{t'})$
\STATE $J^{n,m}=f_{\text{cost}}(X^{n,m}, U^n)$
\ENDFOR
\STATE $J^n=\frac{1}{M}\sum_{m=1}^M f(\frac{J^{n,m}-\min J^{n,m}}{\max J^{n,m} - \min J^{n,m}})$
\ENDFOR
\STATE $\Theta^{k+1}_{t'}=f_{\text{update}}(\{J^n, U^n\}_{n=1}^N)$
\ENDFOR
\STATE Execute $q^K(u_{t',0})$
\STATE $\Theta^0_{t'+1}=f_{\text{recede}}(\Theta^K_{t'})$
\ENDFOR
\end{algorithmic}
\label{alg:Tsallis_VI_MPC}
\end{algorithm}

With the update laws in \cref{sec:update_law} and a choice of the optimality likelihood function $f$, we propose the novel Tsallis \ac{VI}-\ac{MPC} algorithm, summarized in \cref{alg:Tsallis_VI_MPC}.

Given the initial state distribution $p(x_0)$ and a prior policy distribution, $N\times M$ initial states are sampled. Given the initial states, $N$ control trajectories are sampled from the policy distribution. For each control trajectory sample, $M$ state trajectories are propagated for a total of $N\times M$ rollouts. The state transitions at each timestep are sampled from the stochastic dynamics $F(x,u,\epsilon)$ where $\epsilon$ corresponds to system stochasticity (zero mean Gaussian $\epsilon \sim \mathcal{N}(0, \sigma^2_\epsilon\bf{I})$ used in this paper). The cost of each trajectory is normalized to $[0, 1]$ for numerical stability and easier tuning. Depending on the choice of policy distribution class, the policy parameters are updated based on the update laws in \cref{sec:update_law}.

\begin{figure*}[t]
    \centering
    \includegraphics[width=0.325\textwidth]{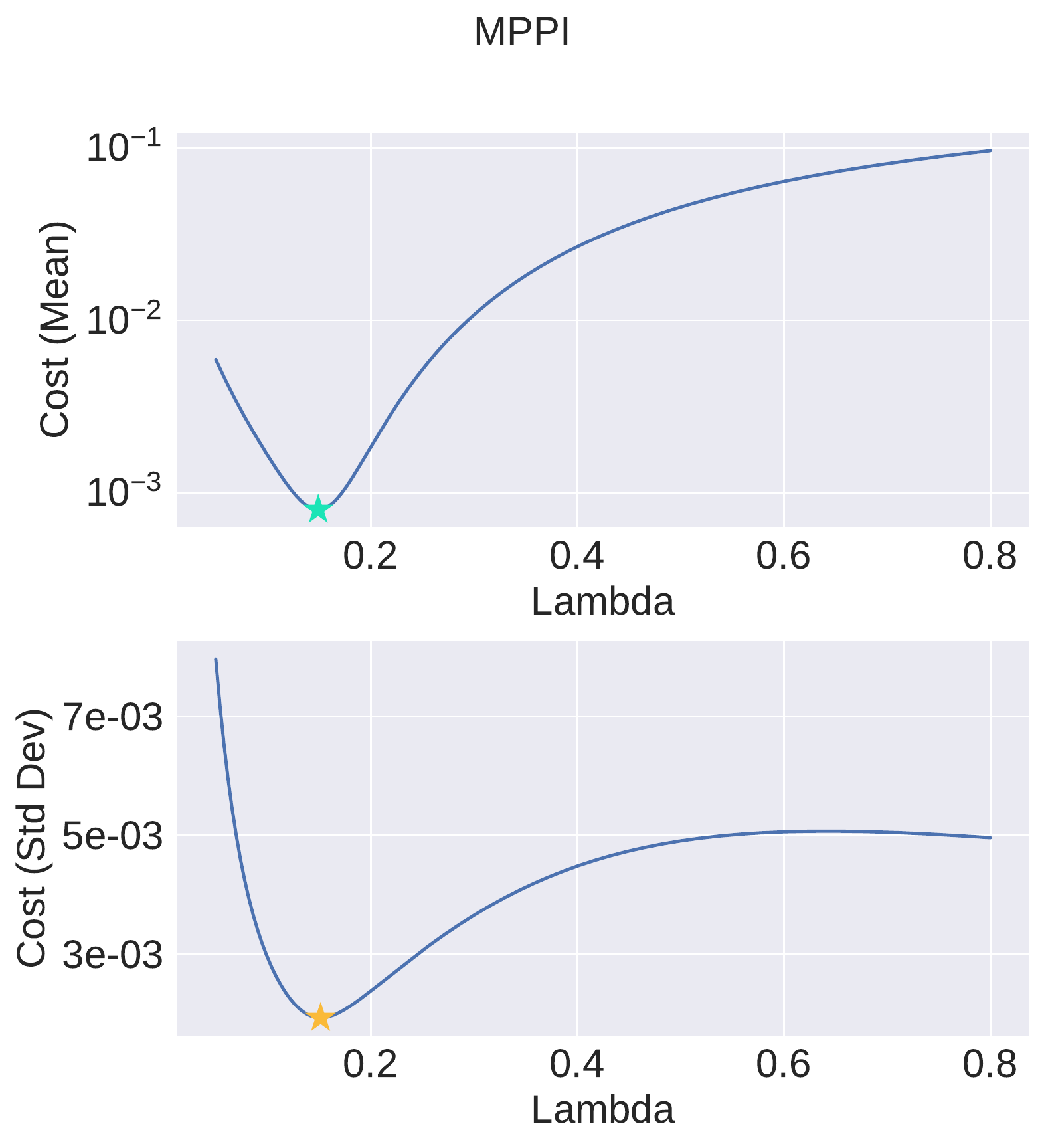}
    \includegraphics[width=0.325\textwidth]{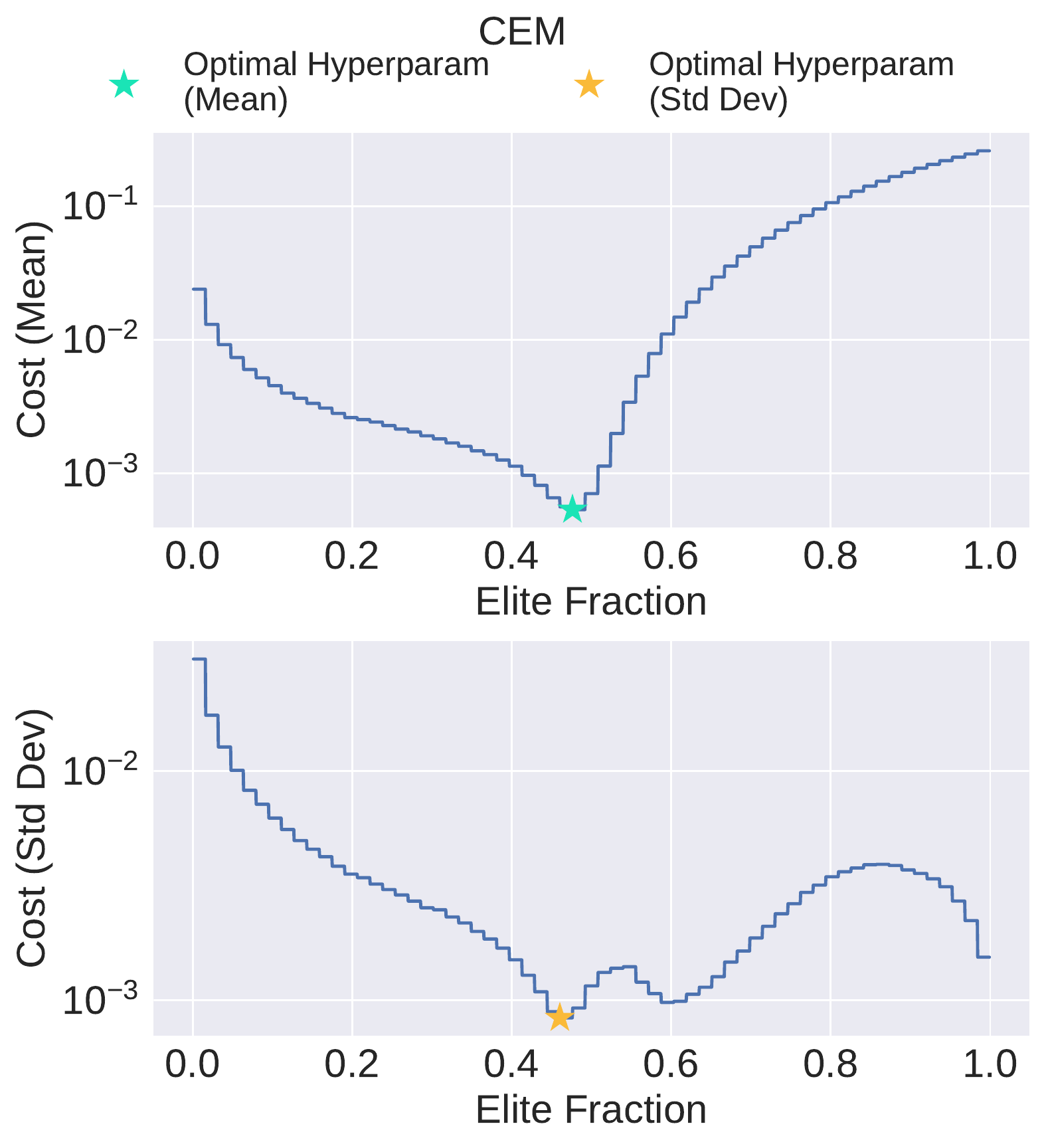}
    \includegraphics[width=0.325\textwidth]{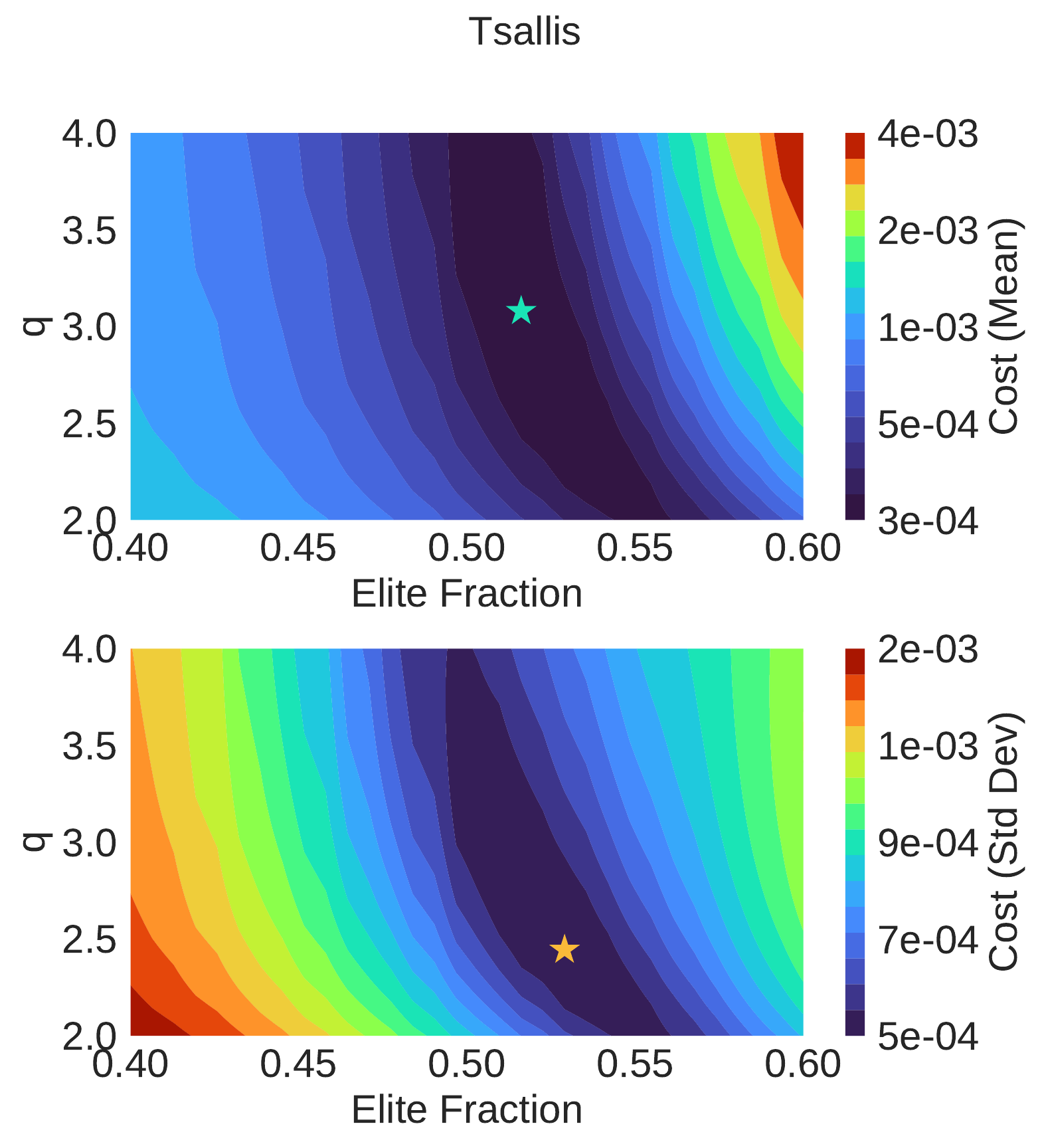}
    \caption{Comparison of mean cost and cost variance for the numerical analysis system \cref{eq:numerical_example_cost}. The cost achieved using the updated $u$ from a single update step, averaged over $4096$ different seeds, is shown over the entire range of the hyperparameter set for \ac{CEM}, \ac{MPPI} and Tsallis VI-SOC.
    The hyperparameters which minimize the mean and standard deviation of the cost are shown as a cyan and orange star respectively.}
    \label{fig:numerical_experiment/hyperparameters}
\end{figure*}
\begin{table}[t]
    \centering
    \caption{Comparison on a simple single stage stochastic optimization problem \cref{eq:numerical_example_cost}. The best values are boldfaced.}
    \begin{tabular}{
        @{}
        l
        S[table-format=1.2e-2,table-number-alignment=right,scientific-notation=true,table-auto-round]
        S[table-format=1.2e-2,table-number-alignment=right,scientific-notation=true,table-auto-round]
        S[table-format=1.2e-2,table-number-alignment=right,scientific-notation=true,table-auto-round]
        @{}
    }
    \toprule
    {Algorithm} & {Cost (Mean)} & {Cost (Std Dev)} & {Mean Control Error} \\
    \midrule
    {Tsallis} & \bfseries 0.000265 & \bfseries 0.000480	& \bfseries 0.022042 \\
    {CEM}     & 0.000536 & 0.000926 & 0.050415 \\
    {MPPI}    & 0.000795 & 0.001157 & 0.043311 \\
    \bottomrule
    \end{tabular}
    \label{tab:numerical_experiment}
\end{table}

\begin{figure}[t]
    \centering
    \includegraphics[width=\linewidth]{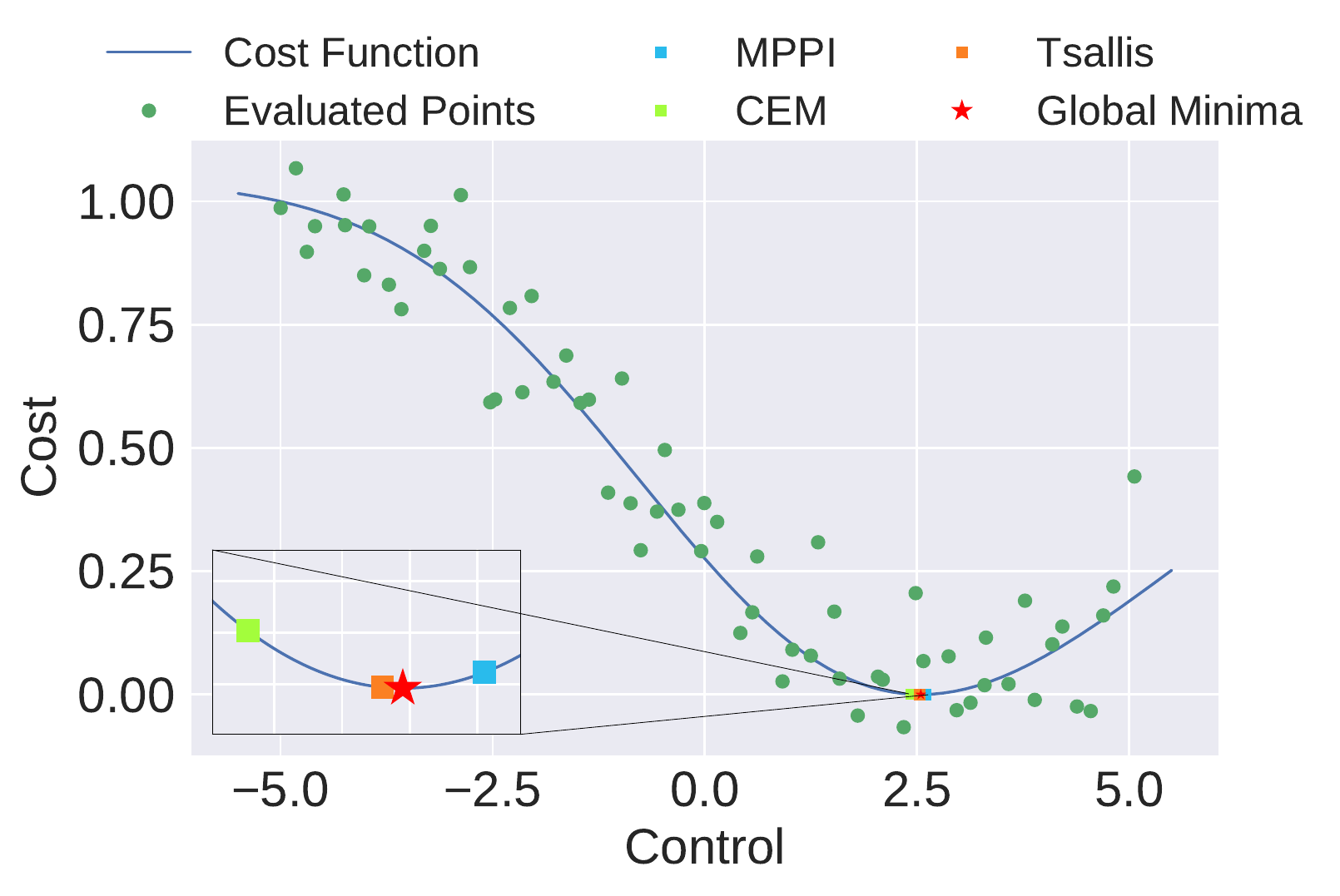}
    \caption{A comparison of the updated means for \ac{CEM}, \ac{MPPI} and Tsallis VI-SOC using the best set of hyperparameters on a single realization of $64$ noisy samples of $u$ (green) for the numerical analysis system.
    Each sample is a noisy realization.
    }
    \label{fig:numerical_experiment/single_realization}
\end{figure}

For the first iteration, we perform additional warm-up iterations by running the optimization loop (line 5 to 14) for a larger $K_\text{warmup}$ iterations before executing the first control and performing $K$ optimization iterations in the ensuing MPC steps.

\textbf{Control Selection:} With the optimized policy distribution from the \ac{VI}-\ac{SOC} framework, the control to be executed on the real system is selected differently based on the choice of policy class. For the unimodal Gaussian policy, the mean of the distribution is used. For the Gaussian mixture policy, the mean of the model with the highest mixture weight is executed. In terms of the Stein policy, the Stein particle with the highest weight is used.

\textbf{Receding Horizon:} After the control execution, the policy distribution is shifted to warm start the optimization at the next \ac{MPC} timestep. Let $\theta'$ be the next iteration's starting sequence and $\theta$ be the current iteration's sequence and set $\theta'_t = \theta_{t+1}$ for $t = 0, ... , T-2$. Finally, set the last item in the new sequence as $\theta'_{T-1} = \theta_{T-1}$.
This is known as the receding horizon technique in \ac{MPC}. Note that in  \cref{alg:Tsallis_VI_MPC}, $\Theta=\{\theta_0, \cdots,\theta_{T-1}\}$.

\section{Simulations}
\label{sec:simulation}
In this section, we compare the proposed Tsallis VI-MPC algorithm against \ac{MPPI} and \ac{CEM}, which represent  state-of-the-art sampling-based \ac{SOC} algorithms.
Since in practice, the shape functions used in \ac{SS}-\ac{SOC} corresponds to the ones which result in \ac{MPPI} and \ac{CEM}, we have chosen to only compare to \ac{MPPI} and \ac{CEM}.
We first validate our analysis on a simple numerical experiment. We then showcase the scalability and performance of the Tsallis VI-MPC algorithm on 2D point mass, quadcopter, ant, manipulator and humanoid systems in simulation under the 3 aforementioned policy distributions.

\subsection{Numerical Analysis}

We verify the analysis in \cref{sec:var_reduction} that Tsallis VI-SOC results in greater variance reduction by comparing the $\exp_r$ cost transform with the \ac{CEM} and \ac{MPPI} cost transforms on the following simple single-stage stochastic optimization problem:
\begin{equation}\label{eq:numerical_example_cost}
    \min_{u} \, \Eb_{\xi} \left[\frac{-\frac{\lambda}{2} e^{\frac{\lambda}{2}(\lambda \sigma^2 - 2u)} \erf\left( \frac{\lambda \sigma^2 - u}{\sqrt{2} \sigma} \right) -c }{d} + 0.1 \xi \right],
\end{equation}
where $\xi \sim \mathcal{N}(0, 1)$, $\erf$ is the corresponding error function, the constants $\lambda$ and $\sigma$ are chosen to be $\lambda=0.2, \sigma=2.5$, and $c$ and $d$ are chosen such that the noiseless objective function is normalized between $0$ and $1$ for $u \in [-5, 5]$. \Cref{fig:numerical_experiment/single_realization} plots the objective function near its minimum and the noisy objective values.

To ensure that only the weight computation of the three methods is tested, we sample $64$ instances of $u$ uniformly from $[-5, 5]$ and use the same set of samples for all three methods.
We use the update law corresponding to the unimodal Gaussian with fixed variance for each algorithm and compare the resulting mean and standard deviation of the cost obtained from each method after a single optimization iteration.
A grid search is performed over $4096$ different hyperparameters, and the optimization metric is computed as the cost averaged over $4096$ random seeds. The results for the best performing hyperparameters are shown in \cref{tab:numerical_experiment} and 
agree with our intuition that the objective function of Tsallis VI-SOC should result in a lower variance compared to \ac{MPPI} and a lower mean compared to \ac{CEM}. In addition, in \cref{fig:numerical_experiment/hyperparameters}, we observe that mean cost increases slower as the elite fraction decreases from the optimal value than when increases, while the opposite is true for cost standard deviation.

\subsection{Controls and Robotics Systems}
The dynamics for the planar navigation and quadcopter tasks are solved via an Euler discretization. Details of the dynamics can be found in \cref{SM:simulation_details}.
For the manipulator, ant, and humanoid tasks, we use the GPU-accelerated Isaac-Gym \cite{macklin2019non} to sample trajectory rollouts in parallel. Additional system stochasticity is injected to each system through the controls channel such that $F(x_t, u_t, \epsilon_t)=F(x_t, u_t+\epsilon_t)$.

The hyperparameters and system configurations for all simulations are included in \cref{SM:simulation_details}. To ensure fair comparisons, all hyperparameters for each method are tuned using a combination of the TPE algorithm \cite{bergstra2011algorithms} from the Neural Network Intelligence (NNI) AutoML framework and hand tuning.

\begin{table*}[t]
    \centering
    \caption{Comparisons of mean and standard deviation of cost against \ac{MPPI} and \ac{CEM} on different systems and policy classes. The policy classes are defined as Unimodal Gaussian (UG), Gaussian Mixture (GM), and Stein (S). Note that the negative of the reward is used for the locomotion tasks (ant and humanoid). The best mean cost and cost variance for each system-policy distribution pair is boldfaced. The mean and standard deviation reduction percentages are included to the right where positive values correspond to a reduction.}
    \begin{tabular}{
        @{}
        l
        c
        S[table-format=6.1,table-number-alignment = right,table-figures-decimal=1,table-auto-round]
        S[table-format=6.1,table-number-alignment = right,table-figures-decimal=1,table-auto-round]
        S[table-format=6.1,table-number-alignment = right,table-figures-decimal=1,table-auto-round]
        S[table-format=6.1,table-number-alignment = right,table-figures-decimal=1,table-auto-round]
        S[table-format=6.1,table-number-alignment = right,table-figures-decimal=1,table-auto-round]
        S[table-format=6.1,table-number-alignment = right,table-figures-decimal=1,table-auto-round]
        S[table-format=6.1,table-number-alignment = right,table-figures-decimal=1,table-auto-round]
        S[table-format=6.1,table-number-alignment = right,table-figures-decimal=1,table-auto-round]
        S[table-format=6.1,table-number-alignment = right,table-figures-decimal=1,table-auto-round]
        S[table-format=6.1,table-number-alignment = right,table-figures-decimal=1,table-auto-round]
        @{}
    }
    \toprule
    & & \multicolumn{2}{c}{MPPI} & \multicolumn{2}{c}{CEM}
    & \multicolumn{2}{c}{Tsallis} & \multicolumn{2}{c}{Tsallis vs MPPI} & \multicolumn{2}{c}{Tsallis vs CEM}\\
    \cmidrule(lr){3-4} \cmidrule(lr){5-6} \cmidrule(lr){7-8} \cmidrule(lr){9-10} \cmidrule(lr){11-12}
    {System} & {Policy} & {Mean} & {Std} & {Mean} & {Std} & {Mean} & {Std} &
    {$\Delta$Mean\%} &
    {$\Delta$Std\%} &
    {$\Delta$Mean\%} &
    {$\Delta$Std\%} \\
    \midrule
    \multirow{3.8}{*}{\begin{tabular}[c]{@{}c@{}}Planar\\Navigation\end{tabular}}
    & \begin{tabular}[c]{@{}c@{}} UG \\ \end{tabular} &
    30023.73 & 3644.07 & 34617.47 & 2523.34 & \bfseries 28714.74 & \bfseries 570.23 & 4.36 & 84.35 & 17.05 & 77.40 \\
    \cmidrule(l){2-12} 
    & \begin{tabular}[c]{@{}c@{}}GM \\ \end{tabular} &
    39313.84 & 8924.76 & 53385.30 & \bfseries 5493.97 & \bfseries 31369.22 & 5683.23 & 20.21 & 36.32 & 41.24 & -3.44 \\
    \cmidrule(l){2-12} 
    & S & 38224.95 & 5169.92 & 38847.56 & 6641.78 & \bfseries 33324.48 & \bfseries 4006.43 & 12.82 & 22.50& 14.22 & 39.68 \\
    \midrule
    \multirow{3.8}{*}{Quadcopter}
    & \begin{tabular}[c]{@{}c@{}}UG \\\end{tabular} &
    15266.15 & \bfseries 1374.04 & 15756.75 & 2707.08 & \bfseries 14673.40 & 1458.36 & 3.88 & -6.14 & 6.88 & 46.13 \\
    \cmidrule(l){2-12} 
    & \begin{tabular}[c]{@{}c@{}}GM \\\end{tabular} &
    22145.51 & 4007.33 & 17654.08 & 1593.86 & \bfseries 16430.56 & \bfseries 1498.24 & 25.81 & 62.61 & 6.93 & 6.0 \\
    \cmidrule(l){2-12} 
    & S &
    29238.25 & 5084.31 & 21064.73 & 2218.02 & \bfseries 15976.02 & \bfseries 1218.32 & 45.36 & 76.04 & 24.16 & 45.07 \\
    \midrule
    \multirow{3.8}{*}{Franka}
    & \begin{tabular}[c]{@{}c@{}}UG \\\end{tabular} &
        64.17 & 8.43 & 66.37 & 18.40 & \bfseries 57.41 & \bfseries 5.53 & 10.6 & 34.5 & 13.6 & 70.1 \\
    \cmidrule(l){2-12} 
    & \begin{tabular}[c]{@{}c@{}}GM \\\end{tabular} &
        67.16 & 8.26 & 68.74 & 13.01 & \bfseries 59.15 & \bfseries 5.37 & 11.98 & 34.9 & 13.90 & 58.5 \\
    \cmidrule(l){2-12} 
    & S & 72.40 & 4.22 & 80.35 & 13.96 & \bfseries 71.77 & \bfseries 3.29 & 0.8 & 21.4 & 10.7 & 76.4 \\
    \midrule
    \multirow{3.8}{*}{Ant}
    & \begin{tabular}[c]{@{}c@{}}UG\\ \end{tabular} &
        -528.7 & 54.1 & -652.6 & \bfseries 26.7 & \bfseries -692.9 & 37.2 & 31.1 & 31.3 & 6.2 & -39.0 \\
    \cmidrule(l){2-12} 
    & \begin{tabular}[c]{@{}c@{}}GM \\ \end{tabular} &
        -621.0 & 55.5 & -659.1 & 53.6 & \bfseries -663.8 & \bfseries 36.3 & 6.9 & 34.7 & 0.7 & 32.4 \\
    \cmidrule(l){2-12} 
    & S &
        -673.7 & 36.1 & -670.0 & 32.2 & \bfseries -677.4 & \bfseries 13.4 & 0.6 & 62.8 & 1.1 & 58.3 \\
    \midrule
    \multirow{3.8}{*}{Humanoid}
    & \begin{tabular}[c]{@{}c@{}}UG\\ \end{tabular} &
        -423.79 & 233.81 & -660.80 & 208.48 & \bfseries -899.27 & \bfseries 80.09 & 112.20 & 65.74 & 36.09 & 61.58 \\
    \cmidrule(l){2-12} 
    & \begin{tabular}[c]{@{}c@{}}GM\\ \end{tabular} &
        -592.85 & 170.15 & -725.92 & 164.92 &\bfseries -738.08 & \bfseries 103.76 & 24.50 & 39.02 & 1.67 & 37.09 \\
    \cmidrule(l){2-12} 
    & S &
        -794.69 & 173.96 & -840.81 & 168.12 & \bfseries -919.67 & \bfseries 105.85 & 15.73 & 39.15 & 9.38 & 37.04 \\
    \bottomrule
    \end{tabular}
    \label{tab:data_table}
\end{table*}

\subsubsection{Planar Navigation}
We first test the different algorithms on a point-mass planar navigation problem. The task is for the point-mass with double-integrator stochastic dynamics to navigate through an obstacle field to reach the target location. The dynamics and obstacle field are set up the same way as \cite{lambert2020stein}. If a crash occurs, a crash cost is incurred and no further movement is allowed.

\subsubsection{Quadcopter}
We also set up a quadcopter 3-D navigation task.
The task is for the quadcopter to reach a target location while avoiding obstacles.
The quadcopter dynamics are taken from \cite{guerra2019flightgoggles}. The quadcopter task is similiar to the planar navigation task, where the system is expected to fly through a randomly generated forest to reach the target location. If a crash occurs, a crash cost is incurred and no further movement is allowed. 

\subsubsection{Franka Manipulator}
\begin{figure} [h]
    \centering
    \includegraphics[width=0.6\linewidth]{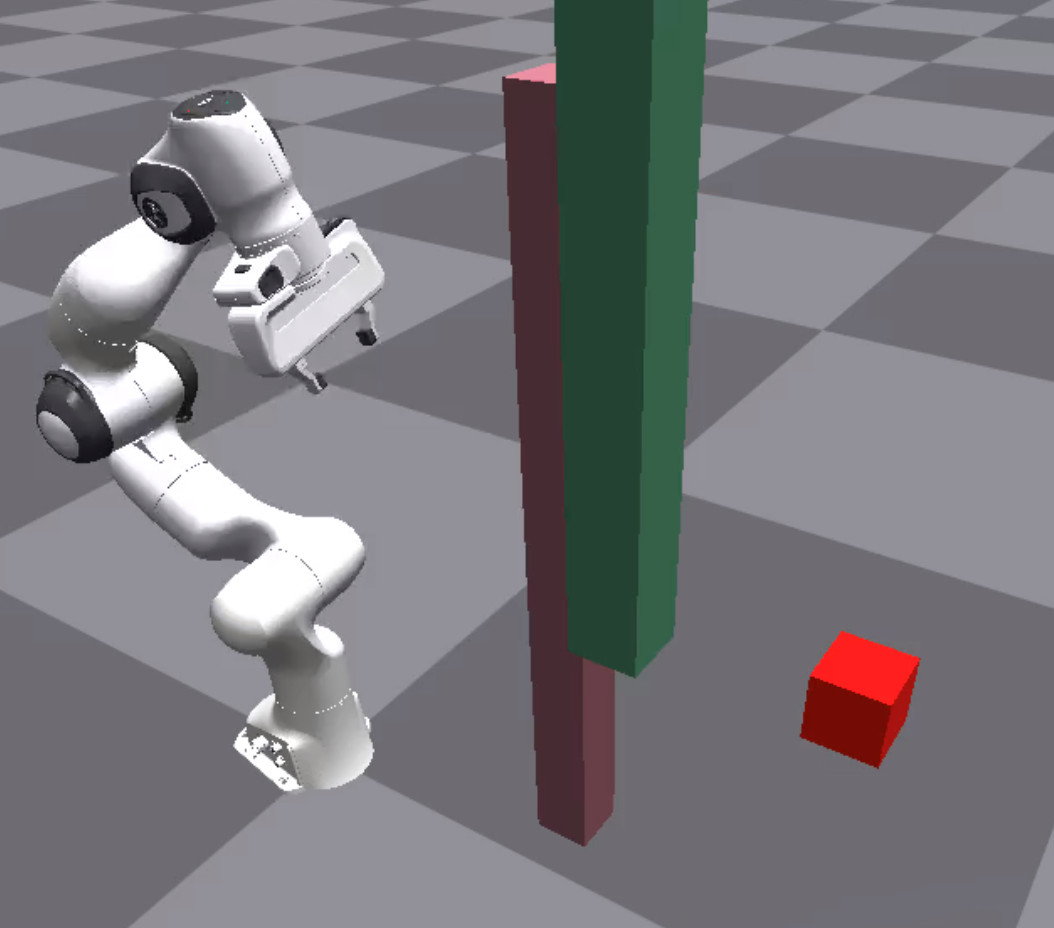}
    \caption{Task setup for Franka manipulator. The goal is to reach the red block while avoiding the pink and green obstacles.}
    \label{fig:franka_task}
\end{figure}

We next test on the Franka manipulator modified to have 7-DOF by removing the last joint and fixing the fingers.
The objective of this task is to move the end effector around the obstacles to the goal. An illustration of the task is in \cref{fig:franka_task}. It is worth noting that no crash cost is in place for the Franka manipulator. Instead, contact is included as a part of the simulation dynamics.

\subsubsection{Ant}
We also consider the task of locomotion. We test on the ant system, which has $29$ state dimensions, $8$ control dimensions, and is a widely used testbed for \ac{RL} algorithms.

\subsubsection{Humanoid}
Finally, we test our approach on the complex humanoid locomotion task. The humanoid system has $56$ state dimensions, $21$ control dimensions, and has very unstable dynamics.

\subsection{Discussion}
In \cref{tab:data_table}, we can see the results of each experiment summarized between each robotic system, policy parameterization, and choice of SOC framework. Across all systems we can see Tsallis VI-MPC results in lower means and variances in most policy parameterizations. The trend continues even as the complexity of the dynamics increases showing that the flexibility provided by the VI-MPC framework is useful even in high dimensional systems.

First in the planar navigation case, we see that Tsallis VI-MPC outperforms the other algorithms in almost all policy parameterizations. 
During testing, there was high variability in the trajectories computed by each algorithm. 
The best performing algorithms would have a large increase in velocity towards the goal state, and would have enough variation in sampled trajectories to ''see`` obstacles via large costs in order to avoid them. When comparing the trajectories computed through the GM policy parameterization we tend to see a reduced ability to stabilize at the goal resulting in larger costs overall when compared to other parameterizations. For these systems, the Stein policy results in longer trajectories to the goal.
Next in the Quadcopter experiments, we observe that the Tsallis VI-MPC outperforms both MPPI and CEM in mean cost for the unimodal Gaussian and Stein policies.
When looking at the high dimensional simulation environments, we see the same trend of Tsallis VI-MPC reducing the mean and variance in comparison to CEM and MPPI holds even for complex systems with contact dynamics.



The performance of the Tsallis VI-MPC can be attributed to the ability of the algorithm to sample policies that are low cost (via the elite fraction), but then improve beyond the CEM by performing the cost weighted averaging similar to MPPI. 
These characteristics of Tsallis VI-MPC are additionally heavily related to the choice of hyperparameters. From hand tuning, we observe that reducing the elite fraction generally results in lower mean cost but higher standard deviation and vice versa. As the cost transform approaches that of CEM or MPPI, the mean and standard deviation approaches their corresponding value.
This verifies that the Tsallis VI-SOC framework can be thought of as an interpolation between CEM and MPPI to a certain degree. The best performing configuration is usually somewhere in between these two extremes.

Finally note that even though the Tsallis VI-SOC framework is typically used over multiple iterations in optimization schemes where the parameters can evolve as the optimization progresses, we have shown the benefits even in MPC mode, where a single iteration of optimization is performed, reiterating the empirical result that a single step has higher reduction in mean cost and variance compared to traditional MPC methods.

\section{Conclusion} 
\label{sec:conclusion}

We present a generalized Variational Inference-Stochastic Optimal Control framework using Tsallis divergence, which allows for additional control of the cost/reward transform and results in lower cost/reward variance. We provide a unifying study of the connections between Tsallis \ac{VI}-\ac{SOC}, \ac{MPPI}, \ac{CEM}, and \ac{SS} methods. The performance and variance reduction benefits of the proposed Tsallis \ac{VI}-\ac{SOC} framework is verified analytically and numerically. We further showcase advantages of the Tsallis \ac{VI}-\ac{MPC} algorithm against MPPI and CEM on 5 different systems with 3 different policy distributions. We leave 2 extensions of this work as future research directions: 1. Implementation of the proposed algorithm on real systems; 2. Comparison against VI-MPC algorithms using other generalized divergences.

\section*{Acknowledgments}
This work is supported by NASA Langley and the NSF-CPS award \#1932288. This work
is supported by Sandia National Laboratories, a multimission laboratory
managed and operated by National Technology and Engineering Solutions of
Sandia, LLC., a wholly owned subsidiary of Honeywell International, Inc., for
the U.S. Department of Energy’s National Nuclear Security Administration
under contract DE-NA-0003525. Additionally, this work is supported by
NASA LaRC.


\bibliographystyle{unsrtnat}
\bibliography{references}




\newpage
\onecolumn
\setcounter{section}{0}
\renewcommand{\thesection}{SM-\arabic{section}} 

\section{Derivation of Optimal Distribution}
\label{SM:optimal_dist_derivation}
To solve for the optimal distribution, we first formulate the Lagrangian as
\begin{align} \label{eq:SM:tsallis_vi_soc:lagrangian}
\mathcal{L}(U,\lambda,\alpha) &= \lambda^{-1} \Eb_{q(U)}[J] + \D{r}{q(U)}{p(U)} \\&+ \alpha \left(1-\int q(U)\rd U\right),
\end{align}
which can be solved by setting $\frac{\partial\mathcal{L}}{\partial q}=0$ as
\begin{align}
\frac{\partial\mathcal{L}}{\partial q}
&= \lambda^{-1} J + \frac{ r}{r-1}\left(\frac{q(U)}{p(U)}\right)^{r-1} - \alpha
=0 \\
\Rightarrow q^*(U)
&= \left(
    \frac{r-1}{ r}\right)^{\frac{1}{r-1}}
    \left(\alpha - \lambda^{-1} J \right)^{\frac{1}{r-1}} p(U) .
\end{align}
Setting $\lambda^{-1} = \alpha(r-1) \tilde{\lambda}^{-1}$, we get
\begin{align}
q^*(U)
&= \left(\frac{\alpha(r-1)}{ r}\right)^{\frac{1}{r-1}}\left(1 - (r-1) \tilde{\lambda}^{-1} J \right)^{\frac{1}{r-1}} p(U) \\
&= \left(\frac{\alpha(r-1)}{ r}\right)^{\frac{1}{r-1}}\expr\left( -\tilde{\lambda}^{-1} J\right) p(U) .
\end{align}
We now use the constraint that $\int q(U) \, \rd U = 1$ by integrating both sides to get
\begin{align}
&1 = \left(\frac{\alpha(r-1)}{ r}\right)^{\frac{1}{r-1}}\int \expr \left( -\tilde{\lambda}^{-1} J \right)p(U)\, \rd U \\
\Rightarrow& \left(\frac{\alpha(r-1)}{ r}\right)^{\frac{1}{r-1}}
= \frac{1}{\int \expr \left( -\tilde{\lambda}^{-1} J\right)p(U)\, \rd U} .
\end{align}
Plugging the normalizing constants back in we can get the optimal policy $q^*$ to be
\begin{equation}\label{eq:SM:tsallis_vi_soc:update_zero_prior}
q^*(U)
=\frac{\expr \left( -\tilde{\lambda}^{-1} J\right)p(U)}
    {\int \expr \left( -\tilde{\lambda}^{-1} J\right)p(U)\, \rd U} .
\end{equation}

\section{Derivation of Update Laws} \label{SM:sec:derive_update_laws}
For each of the following update laws, we make use of the following equality:
\begin{align} \label{eq:derive:kl_mle}
    \pi^{k+1}(U)
    &= \argmin_{\pi(U) \in \Pi} \KL{\tilde{q}^*(U)}{\pi(U)} \\
    &= \argmin_{\pi(U) \in \Pi} \int \tilde{q}^*(U) \log \tilde{q}^*(U) \rd U - \int \tilde{q}^*(U) \log \pi(U) \rd U .
\end{align}
We can drop the first term as it doesn't relate to the $\argmin_{\pi(U) \in \Pi}$:
\begin{align}
    &= \argmin_{\pi(U) \in \Pi}  - \int \tilde{q}^*(U) \log \pi(U) \rd U \\ 
    &= \argmax_{\pi(U) \in \Pi} \int \tilde{q}^*(U) \log \pi(U) \rd U \\ 
    &= \argmax_{\pi(U) \in \Pi} \sum_{n=1}^N w^{(n)} \log \pi(U^{(n)}) .
\end{align}

\subsection{Unimodal Gaussian} \label{SM:subsec:derive_unimodal_gaussian}
For the case of a policy class of unimodal Gaussian distributions where the controls $u_t \in \Rb^{n_u}$ of each timestep are independent, the p.d.f. for each timestep's control has the form
\begin{equation}
    \mathcal{N}(u_t; \mu_t, \Sigma_t)
    = \frac{1}{\sqrt{(2 \pi)^{n_u} \abs{\Sigma_t}}} \exp \left( -\frac{1}{2} (u_t - \mu_t)\T \Sigma_t^{-1} (u_t - \mu_t) \right).
\end{equation}
where we have that 
\begin{align}
    \nabla_{\mu_t} \log \mathcal{N}(u_t; \mu_t, \Sigma_t)
        &= (u_t - \mu_t)\T \Sigma_t^{-1} \label{eq:derive_update:unimodal:mu} \\
    \nabla_{\Sigma_t^{-1}} \log \mathcal{N}(u_t; \mu_t, \Sigma_t)
        &= \frac{1}{2} \Sigma_t - \frac{1}{2} (u_t - \mu_t)(u_t - \mu_t)\T \label{eq:derive_update:unimodal:sigma} .
\end{align}
Minimizing the KL divergence between $\pi(U)$ and $\tilde{q}^*(U)$ and using \cref{eq:derive:kl_mle}, we have
\begin{align}
    \pi^{k+1}(U)
    &= \argmin_{\pi(U) \in \Pi} \KL{\tilde{q}^*(U)}{\pi(U)} \\
    &= \argmax_{\mu, \Sigma} \sum_{n=1}^N \sum_{t=0}^{T-1} w^{(n)} \log \mathcal{N}(U^{(n)}; \mu_t, \Sigma_t).
\end{align}
Hence, we have that
\begin{align}
     0 &= \sum_{n=1}^N w^{(n)} \nabla_{\mu_t} \log \mathcal{N}(U^{(n)}; \mu_t, \Sigma_t) \\
     0 &= \sum_{n=1}^N w^{(n)} \nabla_{\Sigma_t} \log \mathcal{N}(u_t^{(n)}; \mu_t, \Sigma_t).
\end{align}
Using \cref{eq:derive_update:unimodal:mu,eq:derive_update:unimodal:sigma} then results in the update laws
\begin{align}
    \mu_t &= \sum_{n=1}^N w^{(n)} u_t^{(n)} \\
    \Sigma_t &= \sum_{n=1}^N w^{(n)} (u_t^{(n)} - \mu_t) (u_t^{(n)} - \mu_t)\T .
\end{align}
\subsection{Mixture of Gaussian} \label{SM:subsec:derive_gmm}
For a Gaussian mixture model $\pi(U; \theta)$ with $L$ components
with parameters $\theta \coloneqq \{ \theta_l \}_{l=1}^{L}$ and $\theta_l \coloneqq \{ (\phi_l, \{ \mu_{l, t}, \Sigma_{l, t} \}_{t=0}^{T-1} )$ such that
\begin{align}
    \pi(U; \theta) &= \prod_{t=0}^{T-1} \pi_t(u_t; \theta) \\
    \pi_t(u_t; \theta) &\coloneqq \sum_{l=1}^L \pi_m \mathcal{N}(u_t; \mu_l, \Sigma_l).
\end{align}
Directly trying to minimize the KL divergence between $\pi(U; \theta)$ and $\tilde{q}^*(U)$ and using \cref{eq:derive:kl_mle} gives us
\begin{align}
    \theta^*
    &= \argmax_\theta \sum_{n=1}^N \sum_{t=0}^{T-1} \left( w^{(n)} \log \sum_{l=1}^L \phi_l \mathcal{N}(u_t^{(n)}; \mu_{l, t}, \Sigma_{l, t}) \right) . \label{eq:gmm_em_argmax}
\end{align}
However, it is not straightforward to solve \cref{eq:gmm_em_argmax} directly.

Instead, we can use the expectation maximization algorithm and introduce the latent variables $\{ Z_n \}_{n=1}^N$, $Z_n \in [L]$ which form a categorical distribution $q(Z)$ over each of the $L$ mixtures and denotes which mixture each sample $U^{(n)}$ was sampled from.
Then, we can write $\log \pi(U; \theta)$ as
\begin{align}
    \log \pi(U; \theta) &= \int q(z) \log \pi(U; \theta) \rd z \\
    &= \int q(z) \log \frac{ \pi(U; \theta)p(z|U; \theta) }{p(z | U; \theta)} \rd z \\
    &= \int q(z) \log \frac{ p(U, z; \theta)}{p(z | U; \theta)} \rd z \\
    &= \int q(z) \log \frac{ p(U, z; \theta)}{q(z)} \rd z - \int q(z) \log \frac{p(z|U; \theta)}{q(z)} \rd z \\
    &= F(q, \theta) + \KL{q}{p} .
\end{align}
where $F(q, \theta)$ is the ELBO, and provides a lower bound for the log likelihood $\log p(U; \theta)$.
Now, to optimize $F(q, \theta)$, we estimate $q(z)$ via $p(z | U; \theta^{k})$, where $\theta^{k}$ is the previous iteration's parameters.
Then, the ELBO becomes
\begin{align}
    F(q, \theta)
    &= \int q(z) \log \frac{p(U, z; \theta)}{q(z)} \rd z \\
    &= \int q(z) \log p(U, z; \theta) \rd z - \int q(z) \log q(z) \rd z \\
    &= \int p(z | U; \theta^{k}) \log p(U, z; \theta) \rd z - \int p(z | U; \theta^k) \log p(z | U; \theta^k) \rd z \\
    &= Q(\theta, \theta^k) + H(z | U) .
\end{align}
Since the second term is a function of $\theta^k$ and thus is not dependent on $\theta$, it suffices to optimize $Q(\theta, \theta^k)$.
To do so, the EM algorithm separates this into two steps:
\begin{enumerate}
    \item E-step: Compute $p(z | U; \theta^k)$.
    \item M-step: Compute $\argmax_\theta Q(\theta, \theta^k)$.
\end{enumerate}
In our case, we have that
\begin{align}
    Q(\theta, \theta^k)
    &= \int p(z | U; \theta^{k}) \log p(U, z; \theta) \rd z \\
    &= \ExP{z \sim p(z | U; \theta^k) }{ \sum_{n=1}^N \sum_{l=1}^L \sum_{t=0}^{T-1} \mathbbm{1}\{Z_n = l\} w^{(n)} \left( \log \phi_l + \log \mathcal{N}(u_t^{(n)} | \mu_{l, t}, \Sigma_{l, t}) \right) } \\
    &= \sum_{n=1}^N \sum_{l=1}^L \sum_{t=0}^{T-1} p(Z_n = l | u_t^{(n)}; \theta^k) w^{(n)} \left( \log \phi_l + \log \mathcal{N}(u_t^{(n)} | \mu_{l, t}, \Sigma_{l, t}) \right) \\
    &= \sum_{n=1}^N \sum_{l=1}^L \sum_{t=0}^{T-1} \eta_l(u_t^{(n)}) w^{(n)} \left( \log \phi_l + \log \mathcal{N}(u_t^{(n)} | \mu_{l, t}, \Sigma_{l, t}) \right).
\end{align}
where we have defined $\eta_l(u_t^{(n)}) \coloneqq p(Z_n = l | u_t^{(n)}; \theta^{k})$ for simplicity.
To compute this, note that
\begin{align}
    p(Z_k = l| u_t^{(n)}; \theta^k)
    &= \frac{p(u | Z_n = l; \theta^k) p(Z_n = l; \theta^{k})}{\sum_{l'=1}^L p(u | Z_n = l'; \theta^{k}) p(Z_n = l'; \theta^{k})} \\
    &= \frac{ \phi_l \mathcal{N}(u_t^{(n)}; \mu_l^{k}, \Sigma_l^{k}) }{\sum_{l'=1}^L \phi_l \mathcal{N}(u^{(n)}; \mu_{l'}^{k}, \Sigma_{l'}^{k}) } .
\end{align}
To solve for the optimal $\theta^*$ in the M-step, we apply the first order conditions to $\mu_{l, t}$ and $\Sigma_{l, t}$ to obtain
\begin{align}
    0 &= \sum_{n=1}^N \eta_l(u_t^{(n)}) w^{(n)} (u_t^{(n)} - \mu_{l, t}^{k+1}) \Sigma_{l, t}^{-1} \\
    \implies \mu_{l, t}^{k+1} &= \frac{1}{N_{l, t}} \sum_{n=1}^N \eta_l(u_t^{(n)}) w^{(n)} u_t^{(n)} .
\end{align}
and
\begin{align}
    0 &= \sum_{n=1}^N \eta_l(u_t^{(n)}) w^{(n)} \left( \frac{1}{2} \Sigma_{l, t}^{k+1} - \frac{1}{2} (u_{l, t} - \mu_{l, t}^{k+1})(u_{l, t} - \mu_{l, t}^{k+1} )\T \right) \\
    \implies \Sigma_{l, t}^{k+1} &= \frac{1}{N_{l, t}} \sum_{n=1}^N \eta_l(u_t^{(n)}) w^{(n)} (u_t^{(n)} - \mu_{l, t}^{k+1}) (u_t^{(n)} - \mu_{l, t}^{k+1})\T . 
\end{align}
where we have defined $N_{l, t} \coloneqq \sum_{n=0}^N \eta_l(u_t^{(n)}) w^{(n)}$ for convenience.
Since $\phi_l^*$ has the constraint $\sum_{l=1}^L \phi_l^* = 1$, we solve for $\phi_l^*$ by applying the Lagrangian multiplier $\lambda$, yielding the following Lagrangian:
\begin{equation}
    L_t =
    \sum_{n=1}^N \sum_{l=1}^L \sum_{t=0}^{T-1} \eta_l(u_t^{(n)}) w^{(n)} \big( \log \phi_l + \log \mathcal{N}(u_t^{(n)} | \mu_{l, t}, \Sigma_{l, t} \big) + \lambda \left(1 - \sum_{l=1}^L \phi_l \right) .
\end{equation}
Applying first order conditions for $\phi_l$ gives
\begin{align}
    0 &= \sum_{n=1}^N \sum_{t=0}^{T-1} \eta_l(u_t^{(n)}) w^{(n)} \frac{1}{\phi_l^{k+1}} - \lambda \\
    \implies \phi_l^{k+1} &= \frac{1}{\lambda} \sum_{n=1}^N \sum_{t=0}^{T-1} \eta_l(u_t^{(n)}) w^{(n)} \\
                   &= \frac{1}{\lambda} \sum_{t=0}^{T-1} N_{l, t} .
\end{align}
To solve for the Lagrange multiplier $\lambda$, we plug in the constraint to get
\begin{equation}
    \lambda = \sum_{l=1}^L \sum_{n=1}^N \sum_{t=0}^{T-1} \eta_l(u_t^{(n)}) w^{(n)} = \sum_{l=1}^L \sum_{t=1}^{T-1} N_{l, t} .
\end{equation}
Hence, we finally get that 
\begin{equation}
    \phi_l = \frac{N_l}{\sum_{l'=1}^L N_{l'}} .
\end{equation}
where we have defined $N_l \coloneqq \sum_{t=0}^{T-1} N_{l, t}$.

\subsection{Non-parametric Policy via SVGD} \label{SM:subsec:derive_stein}
We now derive the update law for the choice of a fully non-parametric policy via SVGD.
SVGD solves the variational inference problem
\begin{equation}
    \pi^* = \argmin_{\pi \in \Pi} \{ \KL{\pi(U)}{q^*(U)} \} .
\end{equation}
over the set $\Pi \coloneqq \{ z | z = T(x) \}$ consisting of all distributions obtained by smooth transforms $T$ of random variables $x$, where $x$ is drawn from some tractable reference distribution.

One can show that the direction of steepest descent $\phi^*$ which maximizes the negative gradient $-\nabla_\epsilon \KL{ q_{[T]} }{ p } \vert_{\epsilon=0}$ in zero-centered balls in the RKHS $\mathcal{H}^d$ has the form
\begin{align}
    \phi^*(\cdot) = \Eb_{U \sim \pi} \left[ k(U, \cdot) \nabla_U \log q^*(U) + \nabla_U k(U, \cdot) \right] .
\end{align}
Hence, one can then compute the optimal distribution $\pi^*$ by iteratively applying $\phi^*$ to some initial distribution $p_0$.
However, given that we only have an empirical approximation $\tilde{q}^*$ of the optimal distribution $q^*$, the gradient of $\log q^*$ cannot be easily computed.
To solve this, following \cite{lambert2020stein}, instead of optimizing directly over the distribution of \textit{controls} $U$,
one can instead optimize over some distribution $g$ of \textit{parameters} $\theta$ of a parametrized policy $\hat{\pi}(U; \theta)$, such that
\begin{equation}
    \pi(U) = \Eb_{\theta \sim g} \left[\hat{\pi}_\theta(U) \right] \quad \text{and} \quad q^*(\theta) = \Eb_{U \sim \hat{\pi}(\cdot; \theta)} \left[ q^*(U) \right] .
\end{equation}
By doing so, we obtain the following new \ac{VI} problem and steepest descent direction respectively:
\begin{align}
    g^* &= \argmin_{\theta \in \Theta} \{ \KL{ g(\theta) }{ q^*(\theta) } \} \\
    \hat{\phi}^*(\cdot) &= \Eb_{\theta \sim g} \Big[ \hat{k}(\theta, \cdot) \nabla_\theta \log \Eb_{U \sim \hat{\pi}(\cdot; \theta)} [q^*(U)] + \nabla_\theta \hat{k}(\theta, \cdot) \Big] ,
\end{align}
As a result, we can now take the gradient of $\log \Eb_{U \sim \hat{\pi}(\cdot; \theta)} [q^*(U)]$ as
\begin{align}
    \nabla_\theta \log \Eb_{U \sim \hat{\pi}(\cdot; \theta)} [q^*(U)]
    &= \frac{\nabla_\theta \Eb_{U \sim \hat{\pi}(\cdot; \theta)} [q^*(U)]}{\Eb_{U \sim \hat{\pi}(\cdot; \theta)} [q^*(U)]} \\
    &= \frac{\int q^*(U) \nabla_\theta \hat{\pi}(U; \theta) \, \rd U }{\Eb_{U \sim \hat{\pi}(\cdot; \theta)} [q^*(U)]} \\
    &= \frac{\Eb_{U \sim \hat{\pi}(\cdot; \theta)} [q^*(U) \nabla_\theta \log \hat{\pi}(U; \theta)]}{\Eb_{U \sim \hat{\pi}(\cdot; \theta)} [q^*(U)]} .
\end{align}
Discretizing the above since we only have a discrete approximation $\tilde{q}^*$ of $q^*$, we get that
\begin{equation}
    \nabla_\theta \log \Eb_{U \sim \hat{\pi}(\cdot; \theta)} [q^*(U)]
    = \frac{\sum_{n=1}^N \tilde{q}^*(U^{(n)}) \nabla_\theta \log \hat{\pi}(U^{(n)}; \theta) }{\sum_{n=1}^N \tilde{q}^*(U^{(n)}) } .
\end{equation}
Hence, by letting $g$ be an empirical distribution $g(\theta) = \sum_{l=1}^L \bm{1}_{\{\theta_l\}}$, we obtain the following update rule for the particles $\{\theta_l\}_{l=1}^L$:
\begin{equation}
    \theta_l^{k+1} = \theta_l^k + \hat{\phi}^*(\theta_l^k) .
\end{equation}

\section{Variance Reduction Analysis}
\label{SM:variance_analysis}

The variance reduction analysis is conducted through the connection between Tsallis VI-SOC, MPPI, CEM, and SS-SOC. With the appropriate selection of the shape function, the update laws of Tsallis VI-SOC, MPPI and CEM can be derived from SS-SOC. Associated with the SS-SOC update laws are their corresponding problem formulation in stochastic optimization. For the reparameterized Tsallis VI-SOC framework, its associated problem formulation in optimization is
\begin{equation}
\theta^* = \argmax_\theta \Eb_{p(u;\theta)} \left[\exp\left(\frac{1}{r-1}\log \left(1-\frac{J}{\gamma}\right)\right)\right] .
\end{equation}
Note that the scenario where $J \geq \gamma$ is ignored since the Taylor series expansion and analysis will be around $0<J<\gamma$. Similarly, the optimization problem formulation corresponding to MPPI is 
\begin{equation}
\theta^* = \argmax_\theta \Eb_{p(u;\theta)}\left[\exp(-\lambda^{-1}J)\right] .
\end{equation}
For CEM, the corresponding problem formulation is
\begin{equation}
\theta^* = \argmax_\theta \Eb_{p(u;\theta)}\left[\bm{1}_{J\leq\gamma}\right] .
\end{equation}

To analyze degree of risk aversion of each problem formulation, we can look at its Taylor series expansion. For a cost likelihood function $c(J_\theta)$ where $J_\theta$ is a stochastic cost term parametrized by the parameter $\theta$, the Taylor series expansion around $\ExP{}{J_\theta} \coloneqq \tilde{J}_\theta$ has the form:
\begin{equation}
    c(J_\theta) = c(\tilde{J}_\theta) + c'(\tilde{J}_\theta) \Big(J_\theta - \tilde{J}_\theta \Big) + \frac{1}{2} c''(\tilde{J}_\theta)\Big( J_\theta - \tilde{J}_\theta \Big)^2 + O(J_\theta^3) .
\end{equation}
Taking the expectation:
\begin{equation}
    \ExP{}{c(J_\theta)} \approx c(\tilde{J}_\theta) + \frac{1}{2} c''(\tilde{J}_\theta)\Var(J_\theta) .
\end{equation}
On the other hand, if we define the certainty equivalent cost ${J_\theta}_{CE}$ such that $c({J_\theta}_{CE}) = \ExP{}{c(J_\theta)}$, we have that
\begin{equation}
    f({J_\theta}_{CE}) = c(\tilde{J}_\theta) + c'(\tilde{J}_\theta) \Big({J_\theta}_{CE} - \tilde{J}_\theta \Big) + O({J_\theta}_{CE}^2) .
\end{equation}
Hence,
\begin{align}
    c'(\tilde{J}_\theta) \Big({J_\theta}_{CE} - \tilde{J}_\theta \Big) \approx \frac{1}{2}c''(\tilde{J}_\theta)\Var(J_\theta) .
\end{align}
Defining $\pi_A \coloneqq \tilde{J}_\theta - {J_\theta}_{CE}$ as the \textit{Absolute Risk Premium} (which is a \textbf{negative} quantity for \textbf{cost} minimization as opposed to a \textbf{positive} quantity for \textbf{utility/reward} maximization), we now have that
\begin{align}
    \pi_A
    &\approx \frac{-1}{2} \frac{c''(\tilde{J}_\theta)}{c'(\tilde{J}_\theta)} \Var(J_\theta) \\
    &= \frac{1}{2} A(\tilde{J}_\theta) \Var(J_\theta) ,
\end{align}
where $A(\cdot)$ is the ARA coefficient, defined by
\begin{equation} \label{eq:SM_ARA}
    A(J) = -\frac{c''(J)}{c'(J)} .
\end{equation}
The absolute risk premium measures the difference between the mean cost and average cost scaled by the cost transform as a function of the cost function variance. Therefore, the ARA coefficient indicates the degree of risk aversion with respect to the cost function variance.

The ARA coefficients for Tsallis VI-SOC and MPPI can be easily computed from \eqref{eq:SM_ARA} as
\begin{align}
A_{\text{Tsallis}}(J) &= -\frac{r-2}{(r-1)(\gamma-J)}\\
A_{\text{MPPI}}(J) &= \frac{1}{\lambda} .
\end{align}
Since CEM's problem formulation includes the non-differentiable indicator function, we cannot directly apply the same analysis.
However, by taking the indicator function as the limit of the sigmoid function $\sigma$
\begin{equation}
    \bm{1}_{\{J \leq \gamma\}} = \lim_{k \to \infty} \sigma(-k(J - \gamma)) ,
\end{equation}
we can state the CEM problem formulation as
\begin{equation}
\theta^* = \argmax_\theta \ExP{p(u;\theta)}{\lim_{k\rightarrow\infty} \sigma(-k(J-\gamma))} .
\end{equation}
With this, the ARA coefficient can be computed as
\begin{equation}
A_{\text{CEM}}(J) = \lim_{k\rightarrow \infty} -k\tanh\left(\frac{1}{2}k(\gamma-J)\right) .
\end{equation}

\section{Simulation Details}
\label{SM:simulation_details}

A table of the system parameters is shown in \cref{tab:system_details}.

\subsection{Planar Navigation}
\begin{figure}[h]
    \centering
    \includegraphics[width=0.3\textwidth]{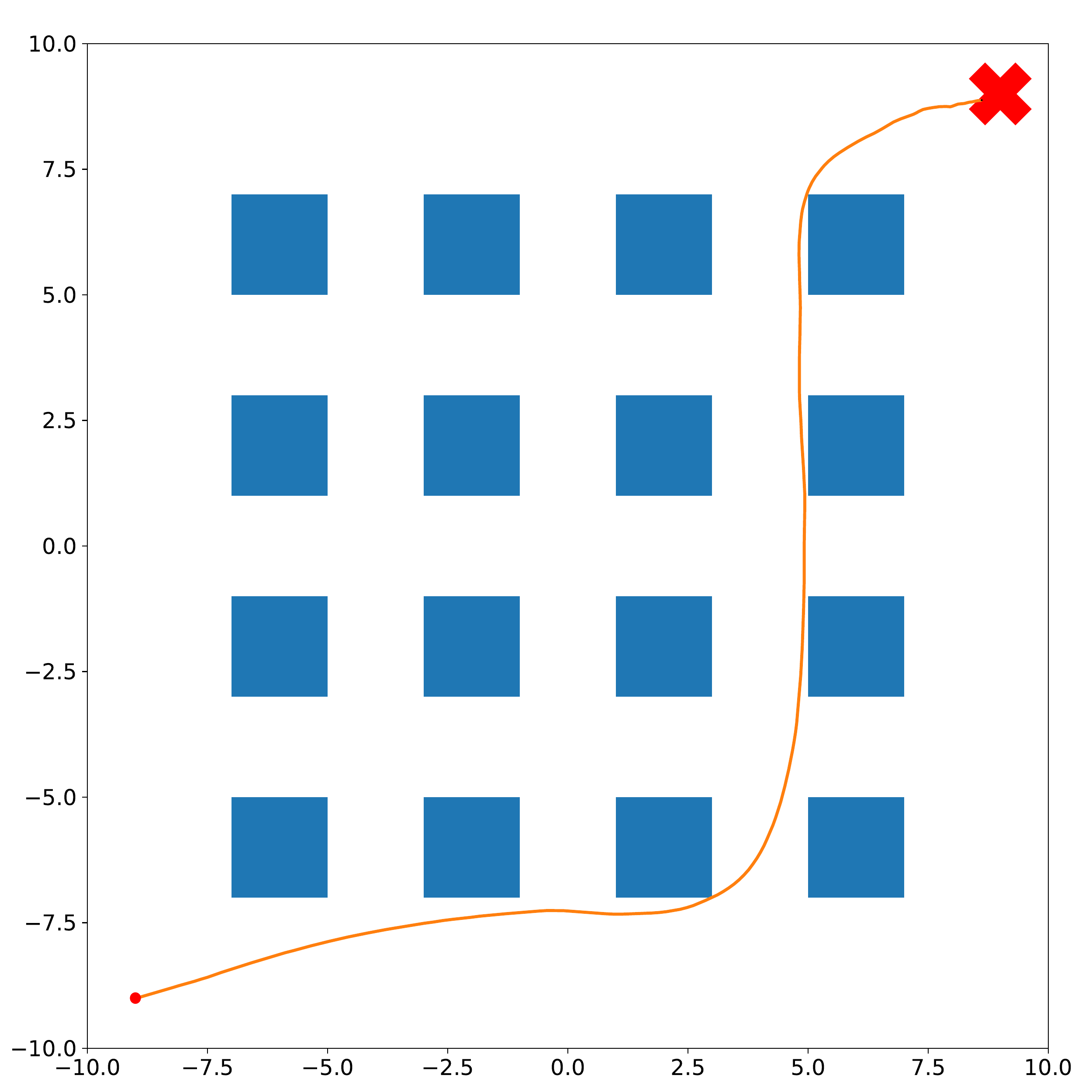}
    \caption{Setup for the Planar Navigation task. The goal is for the robot (orange dot) to reach the goal location (red cross) while avoiding the obstacles (blue squares) in the middle.
    A crash cost of $10000$ is incurred for crashing into the obstacles.}
    \label{fig:planar_task}
\end{figure}

We consider the task of navigating a $2D$ double integrator through an obstacle field (see \cref{fig:planar_task}).
The state consists of $(x_t, \dot{x}_t)$, while the control is $u_t = \ddot{x_t}$, where $x_t, \dot{x}_t, \ddot{x}_t \in \Rb^2$ represent the position, velocity and acceleration of the system.
For the initial position $x_0 \coloneqq (\begin{bmatrix}-9 & -9 \end{bmatrix}\T, \begin{bmatrix}0 & 0\end{bmatrix}\T)$ and goal position $x_{goal} \coloneqq (\begin{bmatrix}9 & 9\end{bmatrix}\T, \begin{bmatrix}0 & 0\end{bmatrix}\T)$, we defined the quadratic cost function:
\begin{equation}
    J(X, U) = 10000 c_{\text{crash}} + (x_T - x_{\text{goal}}) \T Q_f (x_T - x_{\text{goal}}) + \sum_{t=0}^{T-1} (x_t - x_{\text{goal}})\T Q_t (x_t -  x_{\text{goal}}) + u_t\T R u_t ,
\end{equation}
where $c_{\text{crash}}$ is the indicator variable for crashing into an obstacle, and 
\begin{align}
    Q_t &\coloneqq \diag(\begin{bmatrix}0.5 & 0.5 & 0.2 & 0.2\end{bmatrix}) \\
    Q_T &\coloneqq \diag(\begin{bmatrix}0.25 & 0.25 & 1 & 1\end{bmatrix}) \\
    R &\coloneqq \diag(\begin{bmatrix}0.01 & 0.01\end{bmatrix}) .
\end{align}

\subsection{Quadrotor}
The quadrotor task includes navigating a 3D obstacle course with 35 randomly placed obstacles while trying to reach a target location and hover. The control of the quadrotor is done through angular rates and a thrust command, we assume a low level tracking controller that is common for flight systems. The state of the quadrotor $x$ is composed of the 3 coordinates $x,y,z$ angular position represented in quaternions and linear and angular velocities for a total of 13 states. The location of target is represented in cartersian coordinates by $x_{target}$ and is located at (25,25,5). The initial location $x_0$ of the quadrotor is at (0,0,5). For this task, we use the following quadratic cost function:
\begin{equation}
    J(X, U) = \sum_{t=0}^{T-1} 40\| x_{target}  - x\| + 10\| \dot{x}\| + 2\|\omega\| + 1e7 c_{crash},
\end{equation}
where $x$ is the cartesian coordinates of the quadrotor, $\dot{x}$ is the linear velocity, $\omega$ is the angular rates, and $c_{crash}$ is the indicator function that is $1$ if the following conditions are met, and $0$ otherwise
\begin{enumerate}
    \item If the vehicle gets within 0.75 of an obstacle.
    \item If the vehicle goes above 10 meters or below 0 meters in the z axis.
\end{enumerate}

\subsection{Franka}
The Franka manipulator is modified to have $7$-DOF by removing the last joint and fixing the fingers in place.
For state $x \coloneqq (e, \dot{e})$ and control $u \coloneqq \dot{q}$, where $e, \dot{e} \in \Rb^3$ are the position and velocity of the end-effector, and $\dot{q} \in \Rb^7$ are the joint velocities, the objective is to reach the goal position $x_{\text{goal}}$ with obstacles between the starting position of the end effector and the goal. For this task, we use the following quadratic cost function:
\begin{equation}
    J(X, U) = \sum_{t=0}^{T-1} 5 \norm{e_t - x_{\text{goal}}} + 0.1 \norm{\dot{e}_t}.
\end{equation}

\subsection{Ant}
For the Ant, we take inspiration from the cost function for the HalfCheetah from \cite{okada2020variational}
and modify the cost function to reward forward velocity.
For the instantaneous x velocity $v_t$, torso height $h_t$ and angle to the vertical axis $\alpha$, the cost function $J$ is defined as
\begin{equation}
    J(X, U) = \sum_{t=0}^{T-1}  - v_t \bm{1}_{\{h_t \leq 0.7\}} \bm{1}_{\{\abs{\alpha} \geq 0.5\}} + 0.005\norm{u_t}^2 .
\end{equation}

\subsection{Humanoid}
Similar to the Ant, we modify the cost function to reward forward velocity only if the height of the agent's root is not too close or far from the ground and the humanoid is oriented upwards.
For the instantaneous x velocity $v_t$, torso height $h_t$ and angle to the vertical axis $\alpha$,
\begin{equation}
    J(X, U) = \sum_{t=0}^{T-1}  -v_t \bm{1}_{\{0.85 \leq h_t \leq 1.4\}} \bm{1}_{\{\abs{\alpha} \geq 0.5\}} + 0.005\norm{u_t}^2 .
\end{equation}

\begin{table}[h]
    \centering
    \caption{System parameters}
    \label{tab:system_details}
    \begin{tabular}{
        @{}
        l
        c
        c
        c
        c
        S[table-format=3.4]
        S[table-format=3.4]
        @{}
    }
    \toprule
    System & \wrapcell{Episode\\Length} & {Trials} & \wrapcell{State\\Space} & \wrapcell{Control\\Space} & \wrapcell{Timestep} & \wrapcell{System\\Noise} \\
     & & & {($x_t \in$)} & {($u_t \in$)} & {($\Delta$t)} & {($\sigma$)} \\
    \midrule
    Planar Navigation & 300 & 100 & $\Rb^4$ & $\Rb^2$ & 0.01 & 1.0 \\
    Quadcopter & 400 & 100 & $\Rb^{13}$ & $\Rb^4$ & 0.015 & 6.67 \\
    Franka & 80 & 8 & $\Rb^{14}$ & $\Rb^{7}$ & 0.005 & 0.5 \\
    Ant & 128 & 8 & $\Rb^{29}$ & $\Rb^{8}$ & 0.05 & 0.005 \\
    Humanoid & 128 & 8 & $\Rb^{55}$ & $\Rb^{21}$ & 0.033 & 0.05 \\
    \bottomrule
    \end{tabular}
\end{table}

\begin{table}[h]
    \centering
    \caption{Optimization Hyperparameters Per Dynamics }
    \label{tab:hyperparameters:gaussian}
    \begin{tabular}{
        @{}
        l
        c
        c
        c
        c
        @{}
    }
    \toprule
    {System} & {Warmup Iterations} & {Iterations per Timestep} & {Number of Samples} & {MPC Horizon $T$} \\
    \midrule
    {Planar Navigation} & 8 & 1 & 256 & 96 \\
    {Quadcopter} & 64 & 1 & 1024 & 150 \\
    {Franka} & 32 & 2 & 256 & 20 \\
    {Ant} & 32 & 4 & 256 & 20 \\
    {Humanoid} & 32 & 4 & 1024 & 20 \\
    \bottomrule
    \end{tabular}
\end{table}

\begin{table}[h]
    \centering
    \caption{Hyperparameters for Unimodal Gaussian}
    \label{tab:hyperparameters:gaussian}
    \begin{tabular}{
        @{}
        l
        l
        S[table-format=3.1,table-number-alignment = right,table-figures-decimal=3,table-auto-round]
        S[table-format=3.1,table-number-alignment = right,table-figures-decimal=3,table-auto-round]
        S[table-format=3.1,table-number-alignment = right,table-figures-decimal=3,table-auto-round]
        S[table-format=3.1,table-number-alignment = right,table-figures-decimal=3,table-auto-round]
        @{}
    }
    \toprule
    {System} & {Optimizer} & \begin{tabular}[c]{@{}c@{}}Control Std Dev\\($\sigma$)\end{tabular} & $\lambda^{-1}$ & {Elite Fraction} & $r$ \\
    \midrule
    \multirow{3}{*}{Planar Navigation}
        & {MPPI} & 18.00 & 0.015 & & \\
        & {CEM} & 7.780 & & 0.098 & \\
        & {Tsallis} & 18.667 & & 0.070 & 1.796 \\
    \midrule
    \multirow{3}{*}{Quadcopter}
        & {MPPI} & 0.934 & 3.970\\
        & {CEM}  & 1.056 & & 0.015\\
        & {Tsallis} & 1.45 & & 0.14 & 6.276 \\
    \midrule
    \multirow{3}{*}{Franka}
        & {MPPI} & 0.71997 & 54.7944 & & \\
        & {CEM} & 0.629429 &  & 0.0377874 & \\
        & {Tsallis} & 0.79025 &  & 0.167412 & 1.9894554 \\
    \midrule
    \multirow{3}{*}{Ant}
        & {MPPI} & 0.13 & 54.0 & & \\
        & {CEM} & 0.5 & & 0.05 & \\
        & {Tsallis} & 0.5362 &  & 0.0439 & 5.1794 \\
    \midrule
    \multirow{3}{*}{Humanoid}
    & {MPPI} & 0.58738 & 19.558983 & & \\
    & {CEM} & 0.378038 & & 0.018686738871561533 & \\
    & {Tsallis} & 0.539589381 & & 0.01776816 & 5.9325814 \\
    \bottomrule
    \end{tabular}
\end{table}

\begin{table}[h]
    \centering
    \caption{Hyperparameters for Gaussian Mixture Model}
    \label{tab:hyperparameters:gmm}
    \begin{tabular}{
        @{}
        l
        l
        c
        S[table-format=3.1,table-number-alignment = right,table-figures-decimal=3,table-auto-round]
        S[table-format=3.1,table-number-alignment = right,table-figures-decimal=3,table-auto-round]
        S[table-format=3.1,table-number-alignment = right,table-figures-decimal=3,table-auto-round]
        @{}
    }
    \toprule
    {System} & {Optimizer} & Number of Mixtures $L$ & $\lambda^{-1}$ & {Elite Fraction} & $r$ \\
    \midrule
    \multirow{3}{*}{Planar Navigation}
        & {MPPI} & 4 & 3.051\\
        & {CEM} & 4 & & 0.77\\
        & {Tsallis} & 4 & & 0.036 & 33.294 \\
    \midrule
    \multirow{3}{*}{Quadcopter}
        & {MPPI} & 4 & 0.650\\
        & {CEM} & 4 & & 0.015\\
        & {Tsallis} & 4 & & 0.010 & 1.040 \\
    \midrule
    \multirow{3}{*}{Franka}
        & {MPPI} & 4 & 37.03089 & & \\
        & {CEM} & 4 &  & 0.0272531 & \\
        & {Tsallis} & 4 &  & 0.45 & 1.15 \\
    \midrule
    \multirow{3}{*}{Ant}
        & {MPPI} & 4 & 16.9428 & & \\
        & {CEM} & 4 & & 0.05 & \\
        & {Tsallis} & 4 &  & 0.024 & 5.0 \\
    \midrule
    \multirow{3}{*}{Humanoid}
    & {MPPI} & 4 & 65.22090433465655 & & \\
    & {CEM} & 4 & & 0.012840730283384233 & \\
    & {Tsallis} & 4 & & 0.036912542032967316 & 6.280917657828722 \\
    \bottomrule
    \end{tabular}
\end{table}

\begin{table}[h]
    \centering
    \caption{Hyperparameters for Stein Policy}
    \label{tab:hyperparameters:stein}
    \begin{tabular}{
        @{}
        l
        l
        c
        S[table-format=3.1,table-number-alignment = right,table-figures-decimal=3,table-auto-round]
        S[table-format=3.1,table-number-alignment = right,table-figures-decimal=3,table-auto-round]
        S[table-format=3.1,table-number-alignment = right,table-figures-decimal=3,table-auto-round]
        S[table-format=3.1,table-number-alignment = right,table-figures-decimal=3,table-auto-round]
        @{}
    }
    \toprule
    {System} & {Optimizer} & Number of Particles $N$ & \begin{tabular}[c]{@{}c@{}}Kernel Bandwidth\\Multiplier\end{tabular} &$\lambda^{-1}$ & {Elite Fraction} & $r$ \\
    \midrule
    \multirow{3}{*}{Planar Navigation}
        & {MPPI} & 8 & 1.03 & 71.2 \\
        & {CEM} & 8 & 0.462 &  & 0.023 \\
        & {Tsallis} & 8 & 9.444 & & 0.382 & 3.420 \\
    \midrule
    \multirow{3}{*}{Quadcopter}
        & {MPPI} & 8 & 2.869 & 3.829\\
        & {CEM} & 8 & 12.902 &  & 0.071\\
        & {Tsallis} & 8 & 2.32 & & 0.047 & 2.077 \\
    \midrule
    \multirow{3}{*}{Franka}
        & {MPPI} & 8 & 2.56864 & 50.1596 & & \\
        & {CEM} & 8 & 11.1881 &  & 0.228733 & \\
        & {Tsallis} & 8 & 7.1 &  & 0.37 & 2.0 \\
    \midrule
    \multirow{3}{*}{Ant}
        & {MPPI} & 8 & 6.682712 & 8.159674 & & \\
        & {CEM} & 8 & 7.1 & & 0.15 & \\
        & {Tsallis} & 8 & 7.1 &  & 0.4 & 2.5 \\
    \midrule
    \multirow{3}{*}{Humanoid}
    & {MPPI} & 16 & 9.157033311343476 & 63.527052042637095 & &   \\
    & {CEM} & 16 & 14.816129955170156 & & 0.021217670912240638 & \\
    & {Tsallis} & 16 & 6.0 & & 0.1 & 3.0 \\
    \bottomrule
    \end{tabular}
\end{table}

\section{Additional Results}
\label{SM:additional_results}
In this section we include the results of different framework's sensitivity with respect to its hyperparpameters on the planar navigation example with fixed variance unimodal Gaussian policy in \cref{tab:sensitity}.
We also include the cumulative cost comparison plots of the simulator systems. All plots in this section show error bars of $\pm 1$ standard deviation of the cost.

\begin{table}[h]
    \centering
    \caption{Sensitivity Results for hyperparameters on planar navigation example with fixed variance unimodal Gaussian policy. Tsallis mean variation results are bolded. Note that Tsallis hyperparameters demonstrate lower cost sensitivity when perturbed from their baseline values.}
    \label{tab:sensitity}
    \begin{tabular}{
        @{}
        l
        S[table-format=3.4]
        S[table-format=3.4]
        S[table-format=3.4]
        S[table-format=3.4]
        S[table-format=3.4]
        S[table-format=3.4]
        S[table-format=3.4]
        S[table-format=3.4]
        @{}
    }
    \toprule
    &  \multicolumn{2}{c}{MPPI} & \multicolumn{2}{c}{CEM} & \multicolumn{4}{c}{Tsallis}   \\
    & \multicolumn{2}{c}{$\lambda$} & \multicolumn{2}{c}{$\gamma$} & \multicolumn{2}{c}{$\gamma$} & \multicolumn{2}{c}{$r$}\\
     \cmidrule(lr){2-3} \cmidrule(lr){4-5} \cmidrule(lr){6-7} \cmidrule(lr){8-9}
    & {Mean} & {Std} & {Mean} & {Std} & {Mean} & {Std} & {Mean} & {Std} \\
    \midrule
    +10\% & 30294.44 & 3499.19 & 35557.07 & 1753.46 & 28806.34 & 531.61 & 28873.70 & 572.90 \\
    Baseline & 30023.73 & 3644.07 & 34617.47 & 2523.34 & 28714.70 & 570.20 & 28714.70 & 570.20 \\
    -10\% & 30143.85 & 4273.51 & 34427.06 & 2654.51 & 28571.64 & 591.37 & 28535.40 & 657.00 \\
    Mean Variation (\%) & 0.65 & 6.65 & 1.08 & -12.66 & \bfseries -0.09 & \bfseries -1.53 & \bfseries -0.04 & \bfseries 7.85 \\
    \bottomrule
    \end{tabular}
\end{table}

\begin{figure}
    \centering
    \includegraphics[width=0.75\linewidth]{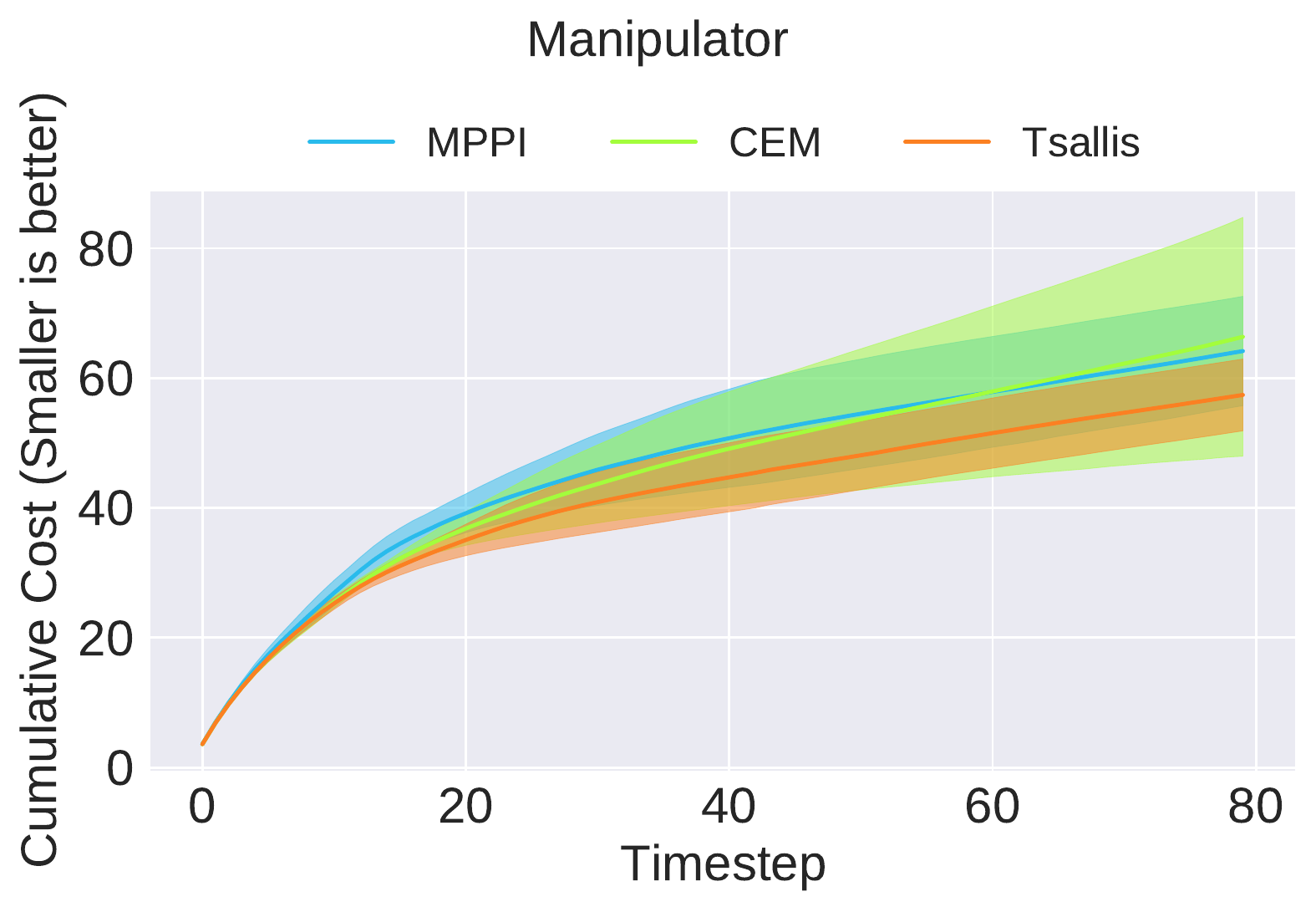}
    \caption{Unimodal Gaussian on Manipulator}
    \label{fig:experiments:franka:fixed_std}
\end{figure}
\begin{figure}
    \centering
    \includegraphics[width=0.75\linewidth]{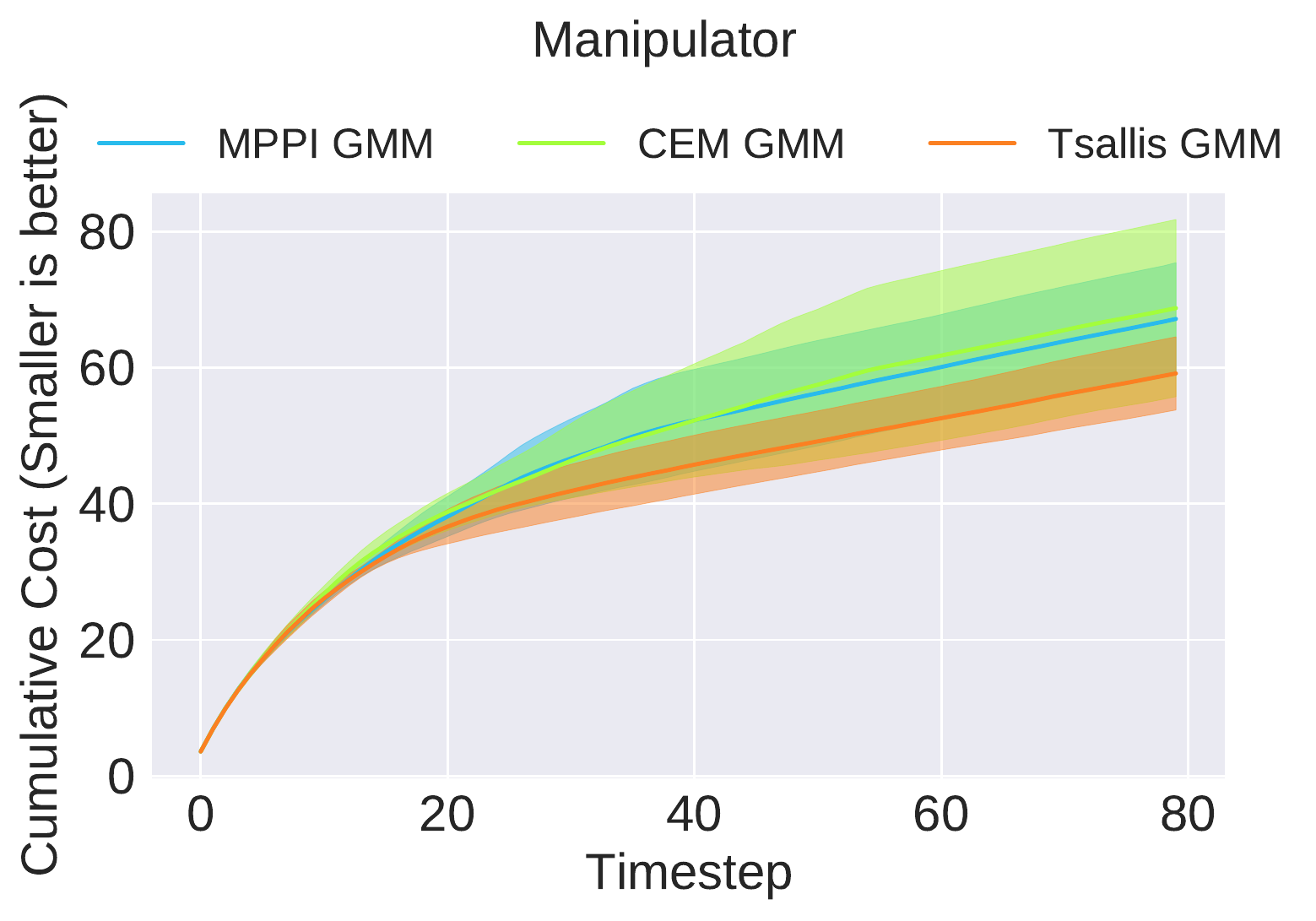}
    \caption{Gaussian Mixture on Manipulator}
    \label{fig:experiments:franka:gmm}
\end{figure}
\begin{figure}
    \centering
    \includegraphics[width=0.75\linewidth]{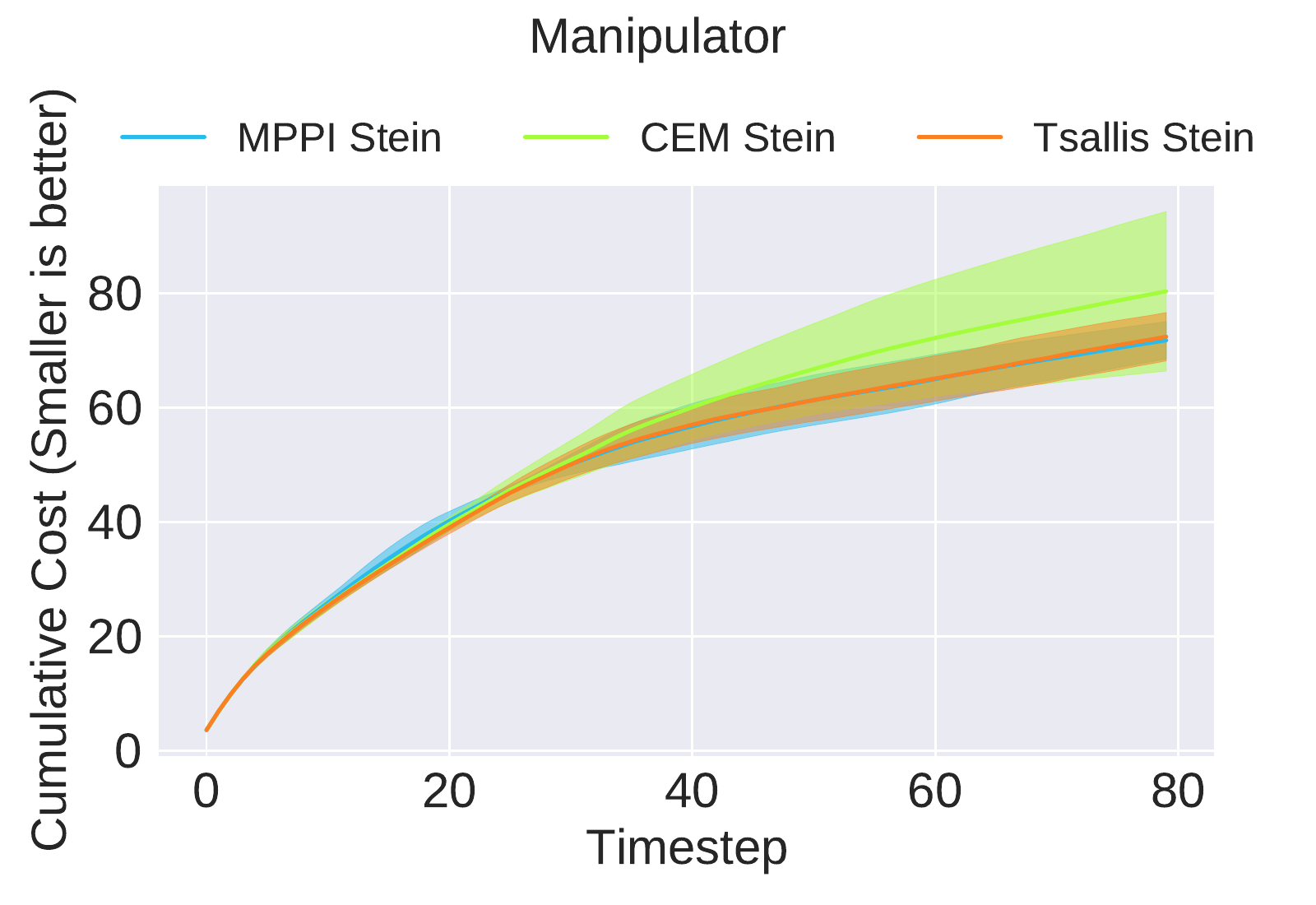}
    \caption{Nonparametric Stein Policy on Manipulator}
    \label{fig:experiments:franka:stein}
\end{figure}
\begin{figure}[h]
    \centering
    \includegraphics[width=0.75\linewidth]{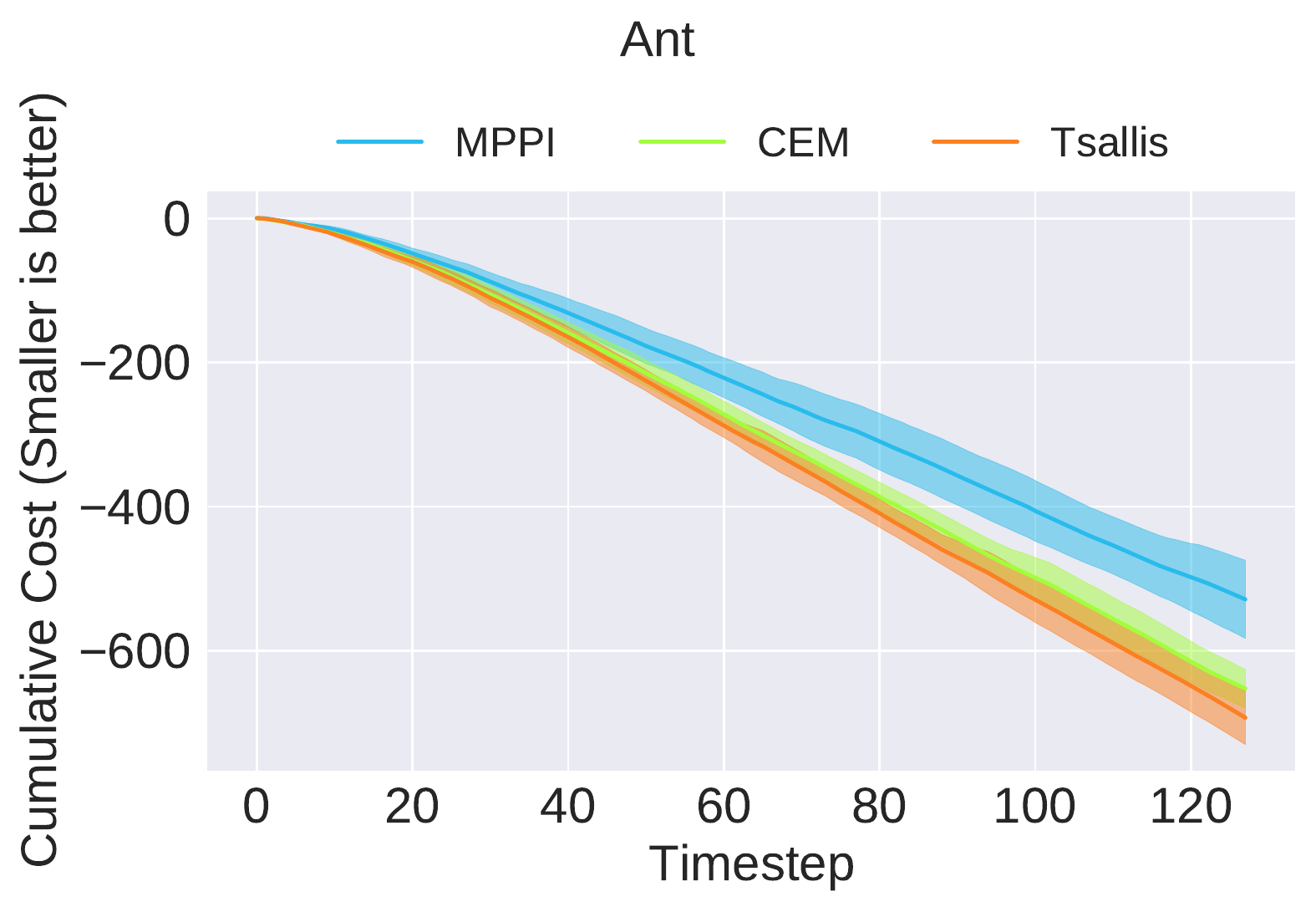}
    \caption{Unimodal Gaussian on Ant}
    \label{fig:experiments:ant:fixed_std}
\end{figure}
\begin{figure}
    \centering
    \includegraphics[width=0.75\linewidth]{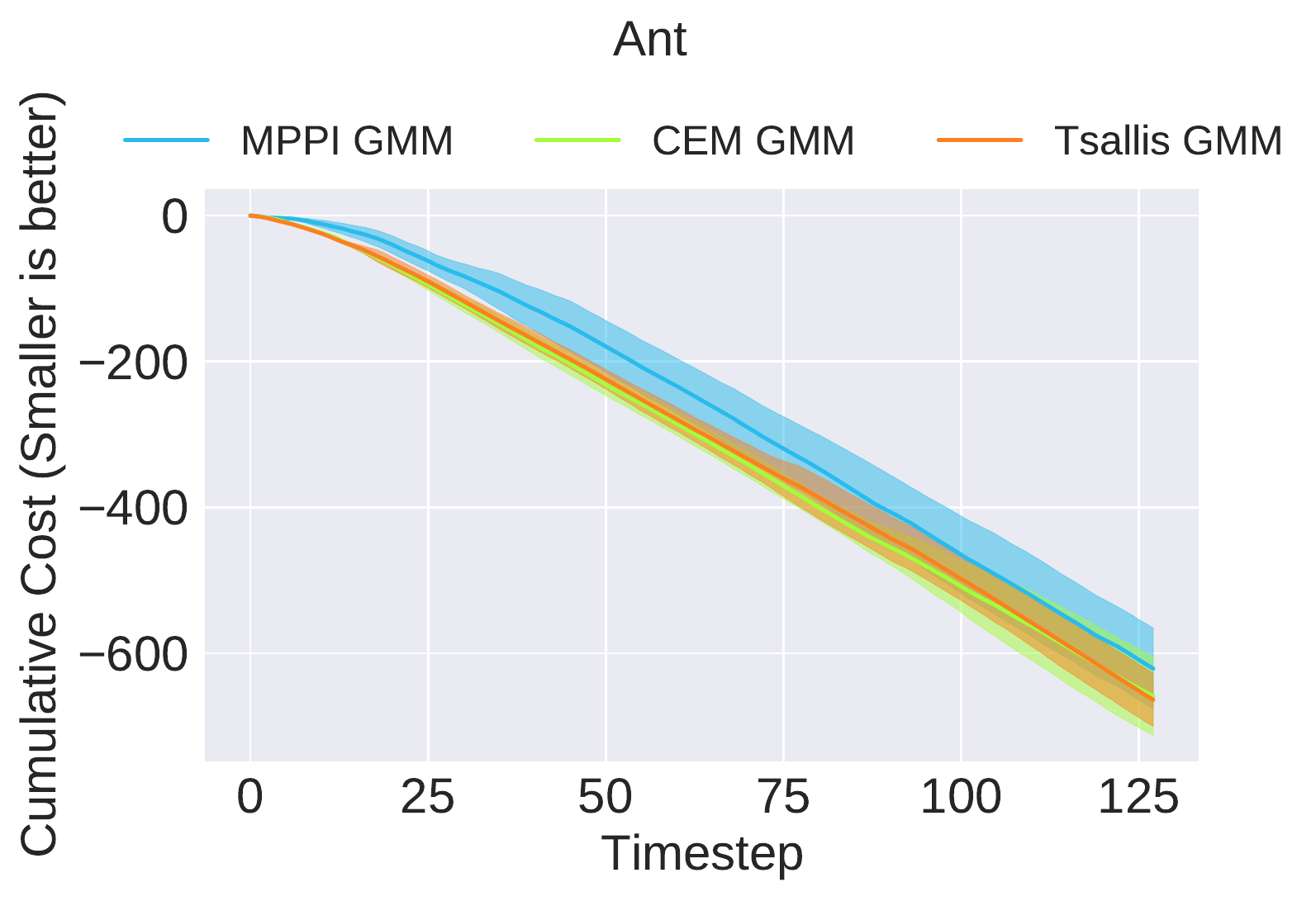}
    \caption{Gaussian Mixture on Ant}
    \label{fig:experiments:ant:gmm}
\end{figure}
\begin{figure}
    \centering
    \includegraphics[width=0.75\linewidth]{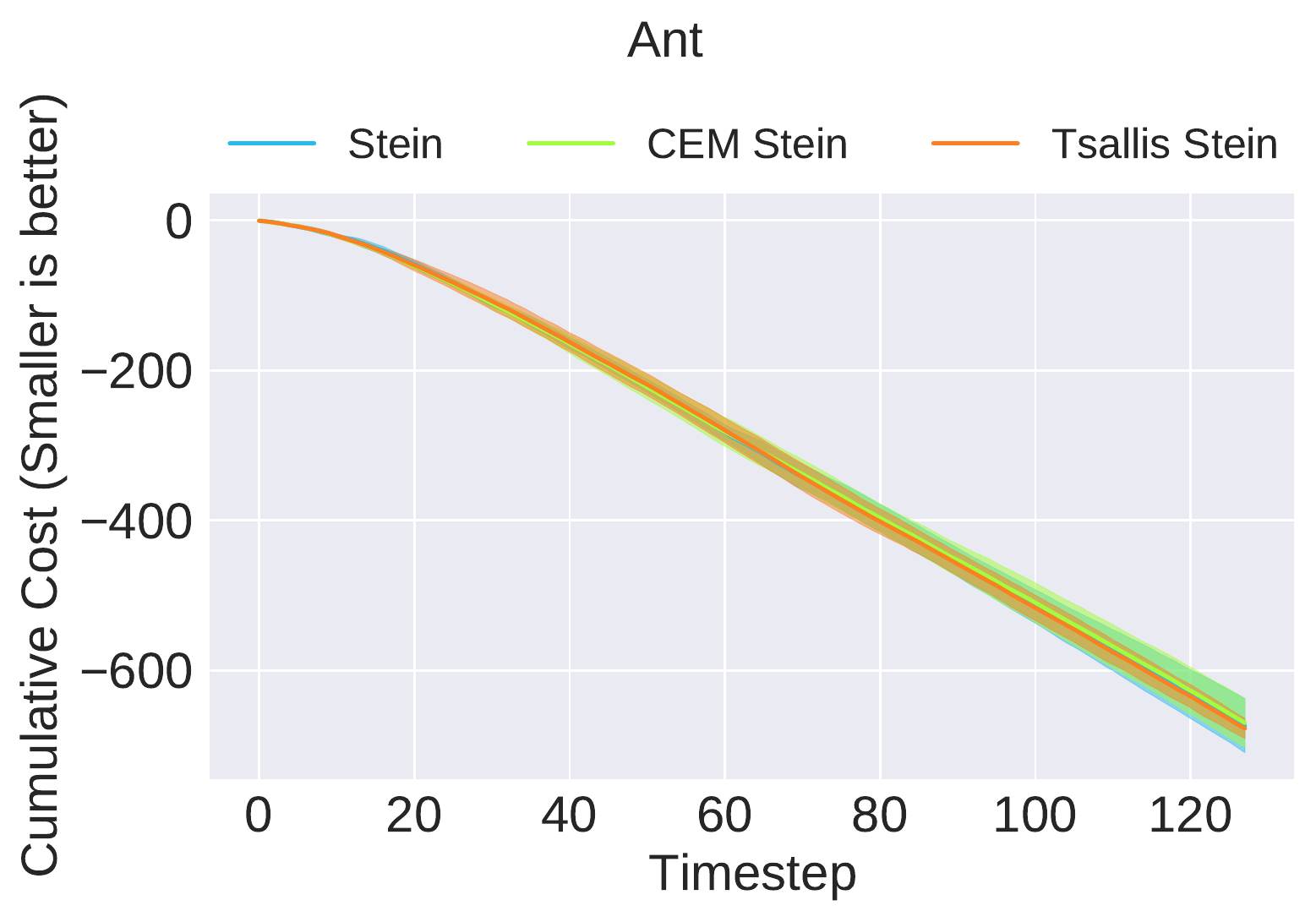}
    \caption{Nonparametric Stein Policy on Ant}
    \label{fig:experiments:ant:stein}
\end{figure}
\begin{figure}
    \centering
    \includegraphics[width=0.75\linewidth]{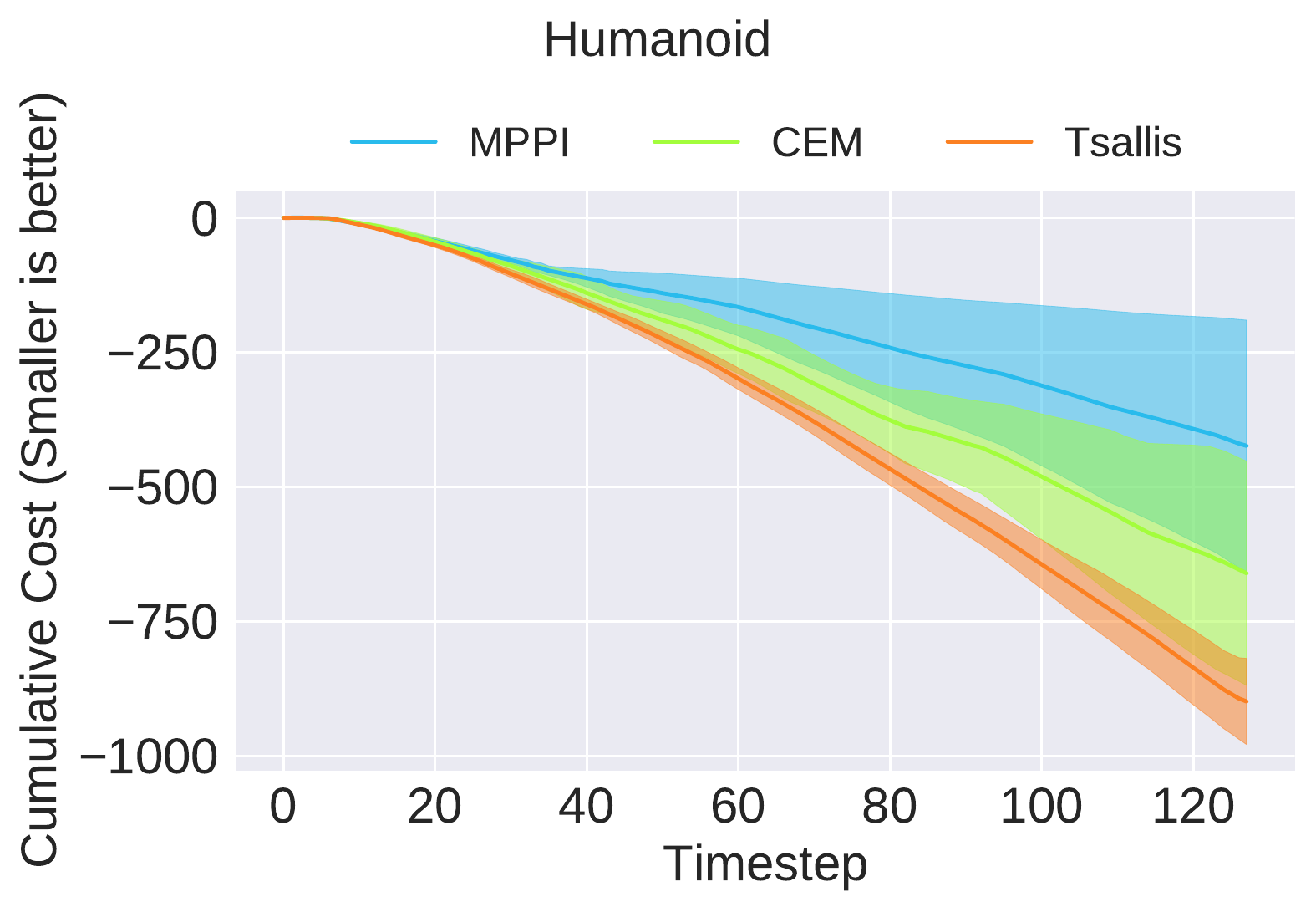}
    \caption{Unimodal Gaussian on Humanoid}
    \label{fig:experiments:humanoid:fixed_std}
\end{figure}
\begin{figure}
    \centering
    \includegraphics[width=0.75\linewidth]{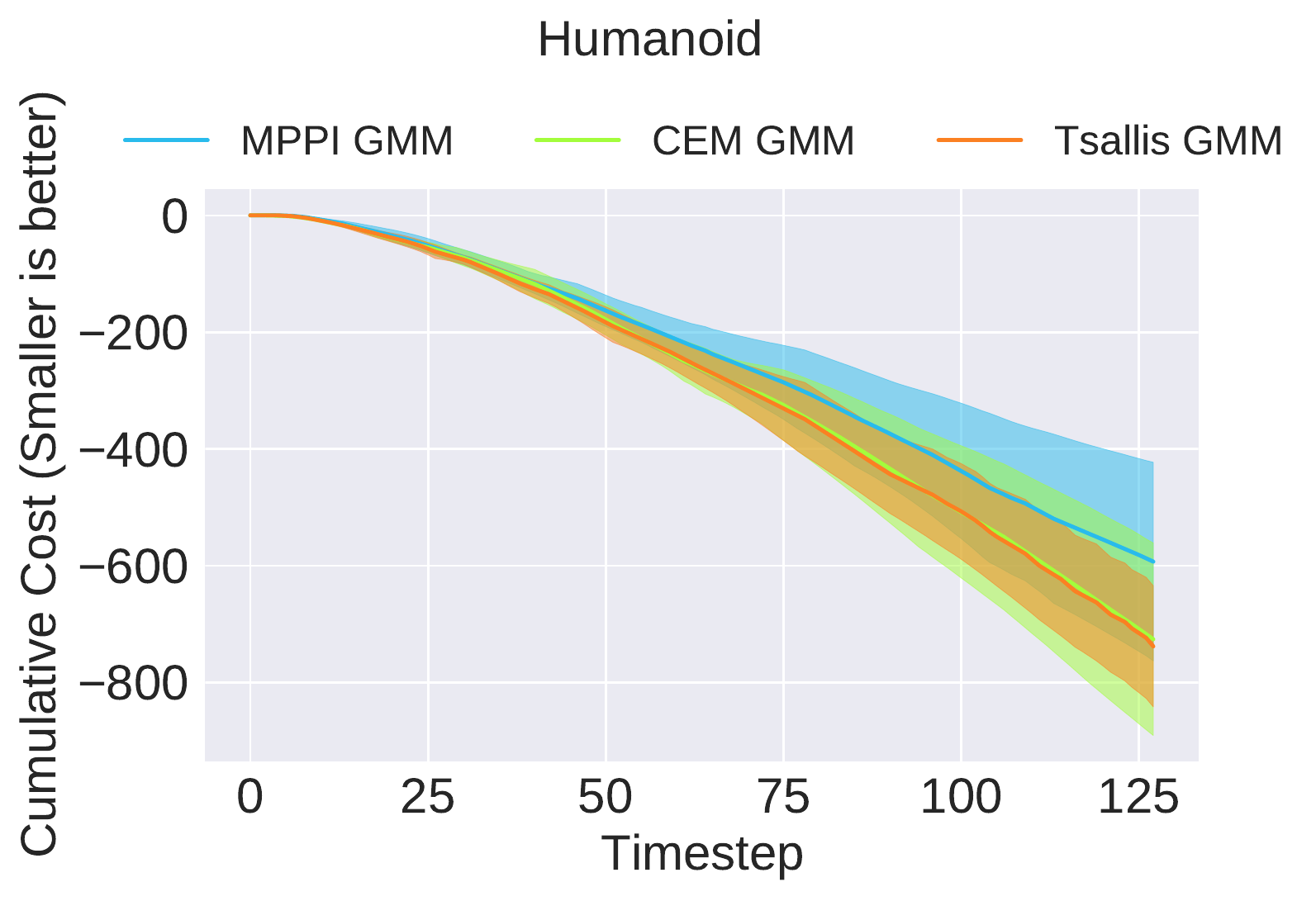}
    \caption{Gaussian Mixture on Humanoid}
    \label{fig:experiments:humanoid:fixed_std}
\end{figure}
\begin{figure}
    \centering
    \includegraphics[width=0.75\linewidth]{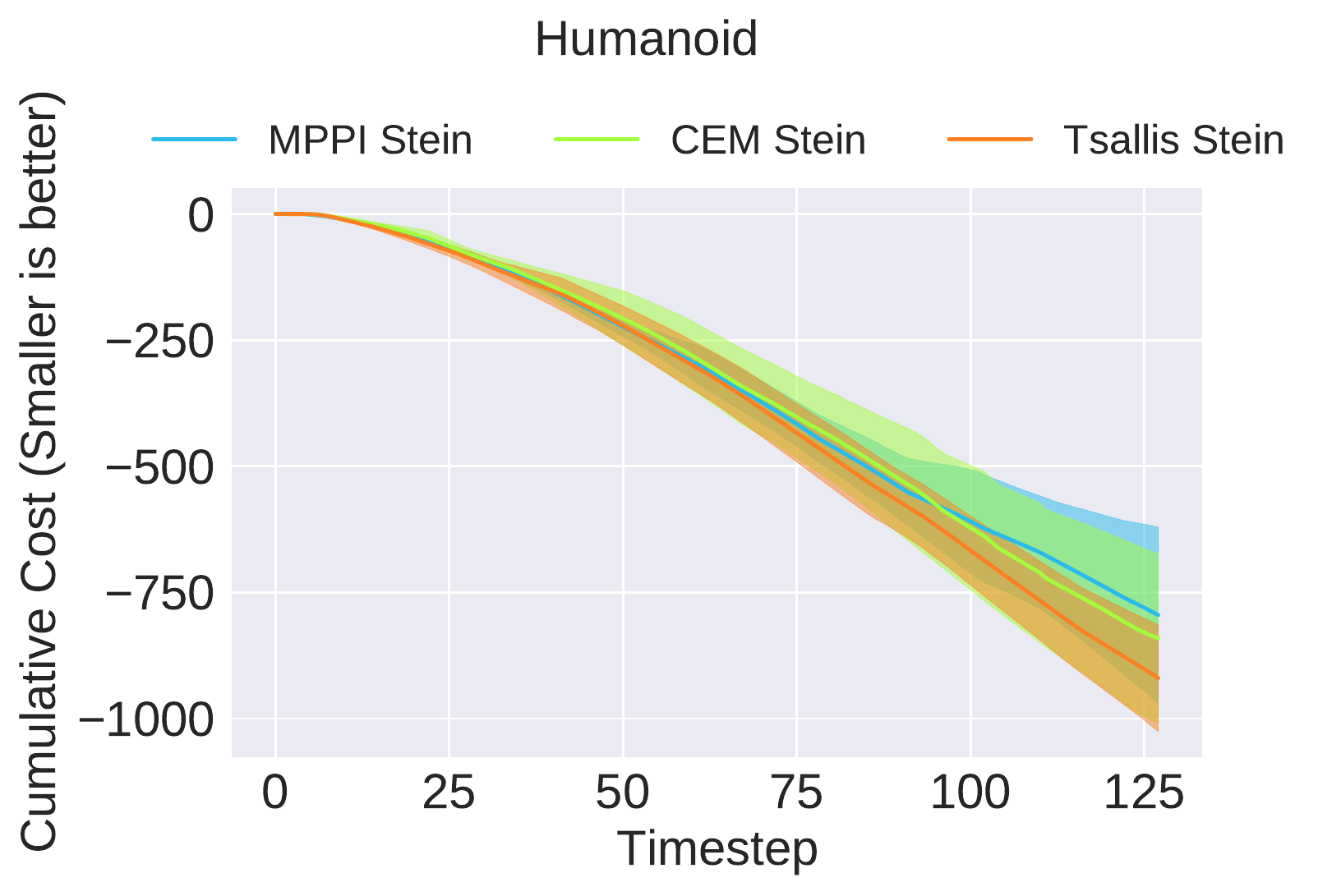}
    \caption{Nonparametric Stein Policy on Humanoid}
    \label{fig:experiments:humanoid:fixed_std}
\end{figure}


\end{document}